\documentclass[journal]{IEEEtran}

\usepackage{times}
\usepackage{epsfig}
\usepackage{graphicx}
\usepackage{amsmath}
\usepackage{amssymb}
\usepackage{multirow}
\usepackage{algorithm}
\usepackage{algorithmic}
\usepackage{subfigure}
\usepackage{mathtools}
\usepackage{threeparttable}
\usepackage{comment}
\usepackage{multirow}
\ifCLASSINFOpdf
\else
\fi
\begin{document}
%
\title{Deep Collective Learning: Learning Optimal Inputs and Weights Jointly in Deep Neural Networks}
%
%
%

\author{Xiang Deng, Zhongfei (Mark) Zhang
\thanks{Xiang Deng is with the Computer Science Department, Watson School, State University of New York at Binghamton, Binghamton, NY 13902, USA (e-mail: xdeng7@binghamton.edu).}
\thanks{Zhongfei Zhang is with the Computer Science Department, Watson School, State University of New York at Binghamton, Binghamton, NY 13902, USA (e-mail: zhongfei@cs.binghamton.edu).}}

\maketitle

\begin{abstract}
It is well observed that in deep learning and computer vision literature, visual data are always represented in a manually designed coding scheme (eg., RGB images are represented as integers ranging from 0 to 255 for each channel) when they are input to an end-to-end deep neural network (DNN) for any learning task.
We boldly question whether the manually designed inputs are good for DNN training for different tasks and study whether the input to a DNN can be optimally learned end-to-end together with learning the weights of the DNN.
In this paper, we propose the paradigm of {\em deep collective learning} which aims to learn the weights of DNNs and the inputs to DNNs simultaneously for given tasks.
We note that collective learning has been implicitly but widely used in natural language processing while it has almost never been studied in computer vision.
Consequently, we propose the lookup vision networks (Lookup-VNets) as a solution to deep collective learning in computer vision.
This is achieved by associating each color in each channel with a vector in lookup tables.
As learning inputs in computer vision has almost never been studied in the existing literature, we explore several aspects of this question through varieties of experiments on image classification tasks.
Experimental results on four benchmark datasets, i.e., CIFAR-10, CIFAR-100, Tiny ImageNet, and ImageNet (ILSVRC2012) have shown several surprising characteristics of Lookup-VNets and have demonstrated the advantages and promise of Lookup-VNets and deep collective learning.
\end{abstract}


 \ifCLASSOPTIONpeerreview
 \begin{center} \bfseries EDICS Category: 3-BBND \end{center}
 \fi
%
\IEEEpeerreviewmaketitle

\section{Introduction}
%
%
%
%
\IEEEPARstart{T}{he} advent of large datasets and computing resources has made deep neural networks (DNNs) as the most popular technology for varieties of applications \cite{krizhevsky2012imagenet, long2015fully, redmon2016you, mobahi2009deep} in computer vision.
One of the most used data types in deep vision models \cite{he2016deep, ren2015faster, liu2016ssd, mistry2016micro, zhang2018multiview, lan2018prior} is images.
It is observed that image pixels are represented by integers as they are almost always coded in discrete color spaces. 
For example, in the RGB space, the pixel values are 8-bit integers (0 to 255).
The integers representing image pixels are first standardized (i.e., a manually designed linear function) and then used as the inputs to DNNs for any learning task.
Thus, the data source of the inputs to DNNs is integers which are involved in the gradient calculation during the training process.
The manually designed inputs have a strong assumption that the color changes in images cause linear value changes in the inputs.
In lights of this, we boldly question whether the manually designed inputs are good for DNN training for different tasks although a DNN with a proper size theoretically can approximate any function \cite{hornik1989multilayer, leshno1993multilayer}.
We study whether the inputs to DNNs can be learned automatically in computer vision.
We take the images in the RGB color space as the examples in this paper, but the idea can be easily extended to the images  (or videos) in other discrete color spaces.
\par


In standard DNNs for computer vision such as VGG \cite{Simonyan15} and ResNet \cite{he2016deep}, only the weights are learned with the RGB inputs during the training process.
Based on the idea of learning, advanced efforts go beyond learning weights and propose to learn the activation function \cite{agostinelli2014learning, molina2019pad}, the pooling function \cite{lee2016generalizing}, and the optimal regularizer \cite{streeter2019learning} by parameterizing them instead of using manually designed functions.
However, to the best of our knowledge, learning the inputs to DNNs in computer vision has never been explored in the existing literature.\par

\begin{figure}[!t]
\centering
\includegraphics[height=2.025cm, width=8.181cm]{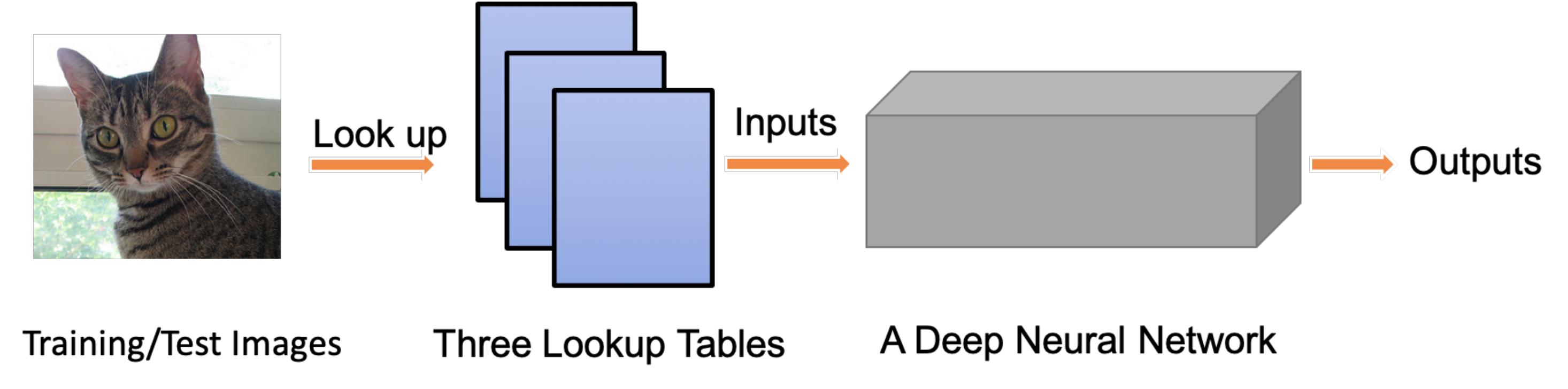}\\
\caption[]{Lookup-VNet}
\label{f1}
\end{figure}

In the paper, we propose {\em deep collective learning} which aims to learn weights in DNNs and inputs to DNNs jointly instead of learning weights alone.
Deep collective learning has been implicitly but successfully used in natural language processing (NLP) where a character or a word is associated with a learned vector \cite{bengio2003neural, ouyang2015sentiment} as the input to a DNN \cite{zhang2018alignment}, but it has not been studied in computer vision.
In light of this, we propose lookup vision networks (Lookup-VNets) as a solution to deep collective learning in computer vision.\par

As shown in Figure \ref{f1}, a Lookup-VNet comprises a DNN and three lookup tables corresponding to the three GRB channels.
The lookup tables are used to parameterize the inputs by associating each color in each channel with a vector.
The pixel colors in images are used as indices to look up the three tables.
The results are fed into the DNN as the inputs to generate the outputs.
The lookup tables are not designed manually but learned jointly with the weights of the DNN.
We propose two kinds of lookup tables for Lookup-VNets according to whether the pixel color space is compressed, i.e., full lookup tables and compressed lookup tables.
Moreover, we introduce three kinds of table learning strategies, i.e., single-task and single-network learning, cross-network learning, and cross-task learning. \par

Lookup-VNets possess several inherent advantages over the standard DNNs. 
First, Lookup-VNets enable DNNs to learn the optimal inputs end-to-end for given tasks.
Second, from the perspective of image coding, Lookup-VNets can be used to learn the optimal image coding scheme automatically for a given criterion.
For example, for the goal of image storage compression with the criterion of accuracy, experimental results show that the pixel color space can be compressed 4096 ($16^3$) times (from 256 $\times$ 256 $\times$ 256 colors to 16 $\times$ 16 $\times$ 16 colors) without accuracy dropping on CIFAR-10, which indicates that the pixel bits can be reduced from 24 (8$\times$3) bits to 12 (4$\times$3) bits under this setting. \par

On the other hand, due to the vacancy of the existing literature on deep collective learning in computer vision, we explore various aspects of this question with Lookup-VNets, such as vector dimensions in lookup tables, pixel color space compression, and table learning strategies, through varieties of experiments on four benchmark datasets, i.e., CIFAR-10 \cite{krizhevsky2009learning}, CIFAR-100 \cite{krizhevsky2009learning}, Tiny ImageNet\footnote{http://tiny-imagenet.herokuapp.com/}, and ImageNet (ILSVRC2012) \cite{deng2009imagenet}.
The experimental results show that Lookup-VNets are able to match the performances of the corresponding standard DNNs on CIFAR-10, CIFAR-100, and Tiny ImageNet while achieving better performances on the large-scale and challenging dataset ImageNet than those of the corresponding standard DNNs, which indicates the superiority of Lookup-VNets on large-scale and challenging datasets and the promise of deep collective learning in computer vision.
We also observe several surprising characteristics of Lookup-VNets: (1) the vector dimensions in lookup tables have no influence on the test performance (generalization ability) of Lookup-VNets; (2) the commonly used color space can be compressed up to 4096 times without accuracy dropping on CIFAR-10, and 3375 times on CIFAR-100 and Tiny ImageNet.

\par

The main contributions of our work can be summarized as follows:
\begin{itemize}
\item We have studied a new question on whether the inputs to deep vision networks can be optimally learned end-to-end and have proposed a new paradigm of {\em deep collective learning} which aims to learn weights in DNNs and inputs to DNNs simultaneously for given tasks.

\item We have proposed a novel framework Lookup-VNets as a solution to deep collective learning in computer vision.
Lookup-VNets make the inputs to DNNs more flexible as an end-to-end solution. 
From the perspective of image coding, Lookup-VNets can be used to learn the optimal image coding scheme for given goals.


\item Due to the lack of the research on collective learning in computer vision, we explore various aspects of this question with Lookup-VNets such as vector dimensions of lookup tables, pixel color space compression, and table learning strategies.
 
\item Through varieties of experiments on four benchmark datasets, we have found several surprising characteristics of Lookup-VNets.
The experimental results have also demonstrated the promise and advantages of Lookup-VNets, which indicates the potential of deep collective learning in computer vision.
\end{itemize}

\section{Related Work}
\label{headings}
In this part, we first review the literature on deep collective learning in NLP.
Then we review the work on learning components of DNNs in computer vision.

\subsection{Deep Collective Learning in NLP}

Deep Collective Learning has been implicitly but successfully used in the area of NLP.
These efforts in NLP mainly associate each character \cite{lee2017fully}, subword unit \cite{sennrich-etal-2016-neural, wu2016google} or word \cite{bengio2003neural, mikolov2013efficient} with a highly dimensional vector in lookup tables.
These vectors are learned for a task, which means the vectors that the lookup tables assign to the characters or words are not designed manually but discovered automatically in the training process of a neural network on a particular task.
Learning lookup tables (embeddings) has been well studied in NLP with large amounts of well known approaches including but not limited to \cite{bengio2003neural, mikolov2013efficient, sennrich-etal-2016-neural, wu2016google, lee2017fully, pennington2014glove, baroni2014don}.
These learned vectors in lookup tables are used as inputs to neural networks (eg., LSTM \cite{hochreiter1997long}) \cite{zhang2015character, conneau2016very, yao2019graph} and are optimized in the training process of a neural network for a given task, which implies the idea of deep collective learning.
However, deep collective learning has never been studied in computer vision.
We propose Lookup-VNets as a solution to this problem in computer vision, which tries to automatically learn representations for pixel colors to replace the fixed integer representations in RGB space.
\par

\subsection{Learning Components of DNNs}
From another perspective, Lookup-VNets can be considered as learning a component (i.e., the input) of a DNN.
Thus, it is also related to the efforts on learning the components of DNNs.
He et al. \cite{he2015delving} and Agostinelli et al. \cite{agostinelli2014learning} propose to learn the activation function end-to-end by parameterizing them.
Lin et al. \cite{lin2014learning}, Zhu et al. \cite{zhu2019empirical}, and Sun et al. \cite{sun2017learning} propose to learn the pooling strategies based on the attention mechanism instead of using the manually designed ones such as average pooling and max pooling.
Streeter et al. \cite{streeter2019learning} propose to learn the optimal regularizer instead of using the manually fine-tuned ones.
However, to the best of our knowledge, learning the input to a DNN has never been explored in the existing literature.

\begin{figure*}[t]
  \begin{center}
\includegraphics[height=4.3cm]{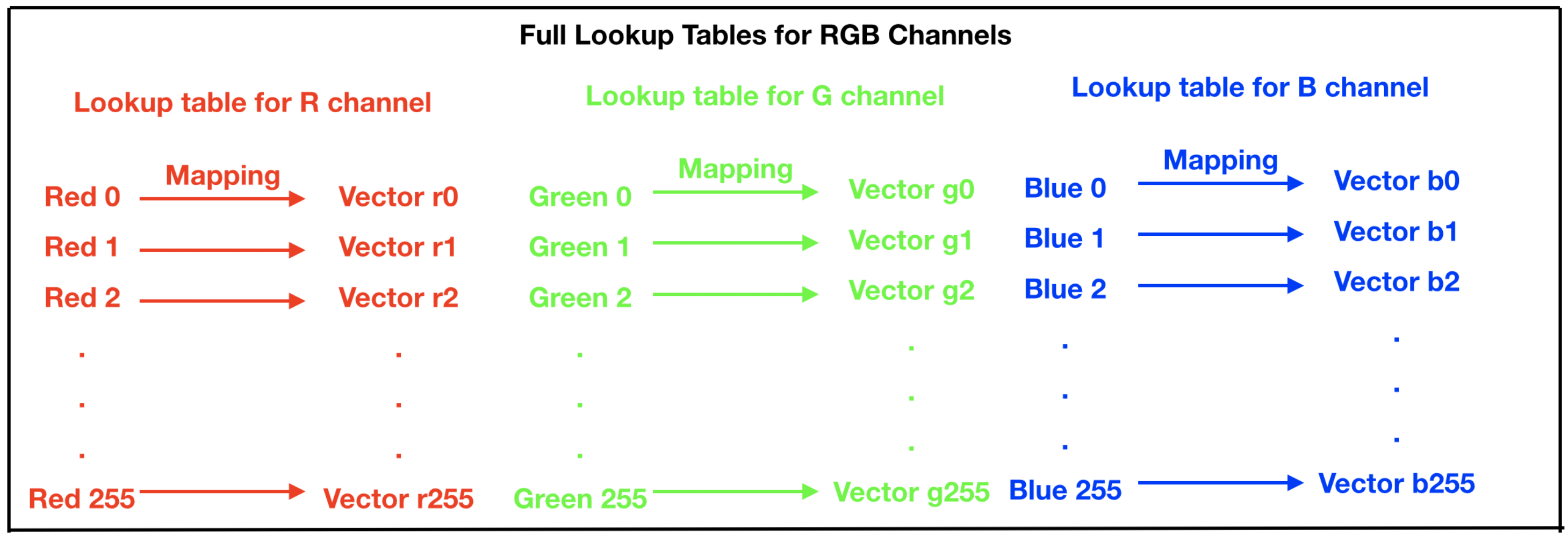} \\
\caption[]{Full Lookup Tables: vectors $r0$ ($g0$, $b0$), $r1$ ($g1$, $b1$), ..., and $r256$ ($g256$, $b256$) are different vectors.}
\label{f2}
\end{center}
\end{figure*}

\section{Our Framework}
To illustrate the connection between standard DNNs and Lookup-VNets, we first review the standard DNNs which only learn the weights with RGB inputs during the training process without computing the gradients with respect to the inputs.
Then we present Lookup-VNets which learn the weights and inputs jointly by associating each color with a vector.
Specifically, we first introduce the full lookup tables and the compressed lookup tables in Lookup VNets; then we present different learning strategies for lookup tables; finally we provide the additional space and computation costs.
We restate that the images in the RGB color space are used as the examples in this paper, but the idea can be easily extended to the images (or videos) in other discrete color spaces.


\subsection{Standard Deep Neural Networks}
Given the training data $(X, Y)$ where $X$ are the images and $Y$ are the targets, the outputs of a DNN $f$ with weights $W$ are $f(X, W)$, and the loss is written as:
\begin{equation}
\label{1}
loss = L(f(X, W), Y)
\end{equation}
where $L(.)$ is any loss function, such as the mean square error or cross entropy. \par

In the training process, the loss function is minimized by the gradient descent based optimizer such as SGD or Adam \cite{kingma2014adam}.
$W$ are updated iteratively based on the gradients on a mini-batch of training samples and the update in step $t$ can be simply expressed as: 
\begin{equation}
\label{2}
W_{t+1} = W_{t} - \lambda \frac{\partial L(f(W_{t}, x_{bat}^{t}), y_{bat}^{t})}{\partial W_{t}}
\end{equation}
where $W_{t}$ are the values of $W$ in step $t$ and $(x_{bat}^{t}, y_{bat}^{t})$ are a mini-batch of training data sampled in step $t$.
As seen from (\ref{2}), only the weights are learned during the training process for a standard DNN.
Lookup-VNets take a further step to update the inputs and weights simultaneously.

\subsection{Lookup-VNets}
\label{lookup}
Different from the standard DNNs, Lookup-VNets learn the weights and inputs jointly instead of learning weights alone.
As shown in Figure \ref{f1}, a Lookup-VNet consists of a DNN and three lookup tables.
The lookup tables associate each color in each channel with a vector. 
According to whether the color space is compressed, we introduce two kinds of lookup tables for Lookup-VNets, i.e., full lookup tables and compressed lookup tables.
We also develop three different strategies for learning lookup tables, i.e., single-network and single-task learning, cross-network learning, and cross-task learning.

\subsubsection{Full Lookup Tables}
The images in the RGB space have three channels, i.e., the red (R) channel, the green (G) channel, and the blue (B) channel.
Each channel has 256 colors represented by 8-bit integers (0 to 255), so the full RGB color space is 256 $\times$ 256 $\times$ 256.
The full lookup tables keep the color space size constant. As shown in Figure \ref{f2}, there are three full lookup tables corresponding to the three RGB channels, and the 256 colors in each channel are associated with 256 distinct vectors.
Thus, the color space is still 256 $\times$ 256 $\times$ 256.
The vector dimension in lookup tables is a hyperparameter, and how it influences the performances is explored in Section \ref{exp}.
\par

\begin{figure*}[ht]
  \begin{center}
\includegraphics[height=5cm]{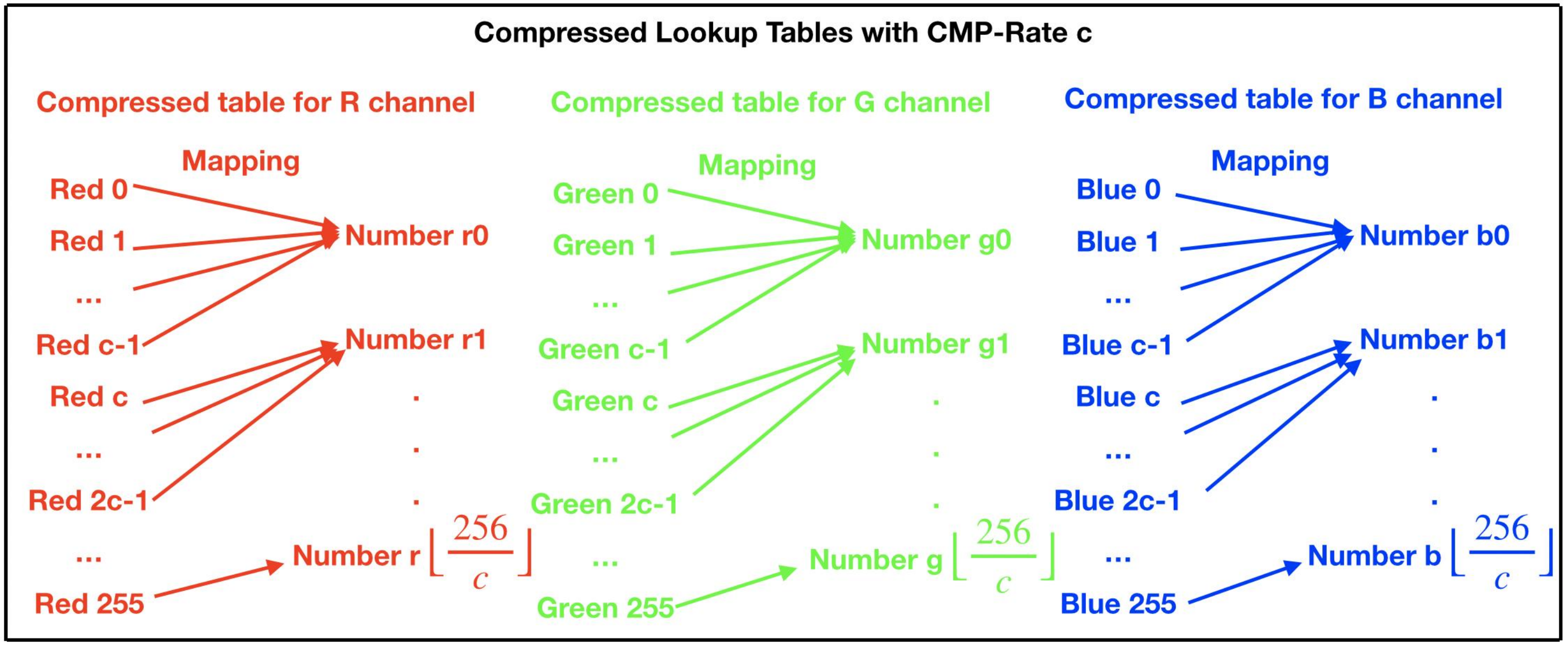} \\
\caption[]{Compressed Lookup Tables with CMP-Rate $c$: $r0$ ($g0$, $b0$), $r1$ ($g1$, $b1$), ..., and $r\lfloor \frac{256}{c} \rfloor$ ($g\lfloor \frac{256}{c} \rfloor$, $b\lfloor \frac{256}{c} \rfloor$) are different numbers.}
\label{f3}
\end{center}
\end{figure*}

\subsubsection{Compressed Lookup Tables}
Full lookup tables map different colors to different vectors so that the color space is still 256$\times$256$\times$256.
We question the necessity of the large color space and propose compressed lookup tables.
The compressed lookup tables compress the color space with a compressing rate (CMP-Rate) $c$.
As shown in Figure \ref{f3}, every $c$ colors in each channel are mapped to a number in the compressed lookup tables, and then there are $\lceil\frac{256}{c}\rceil$ colors in each channel.
Therefore, the whole RGB color space is totally compressed about $c^3$ times, i.e., from 256$\times$256$\times$256 to $\lceil\frac{256}{c}\rceil$$\times$$\lceil\frac{256}{c}\rceil$$\times$$\lceil\frac{256}{c}\rceil$.
It is worth noting that in the compressed tables, every $c$ colors in a channel are mapped to a number not a vector for the goal of compression.
We study how the CMP-Rate $c$ affects the performances of Lookup-VNets in Section \ref{exp}.
An obvious advantage of compressed lookup-tables is that they can be used to save image storage space as each pixel is represented with less bits.

\subsubsection{Single-Task and Single-Network Learning}

In this part, we show how the lookup tables are learned in the scenario of single task and single network.
For an RGB image $x$ with size $m$$\times$$n$$\times$$3$ where $m$, $n$, and 3 are the height, the width, and the channel number, respectively, we use the pixel colors in each channel as indices to look up each lookup table and obtain $x'$:
\begin{equation}
\label{3}
x' = lookup(x,T)
\end{equation}
where $lookup(x,T)$ denotes the result obtained from using the pixels in image $x$ as indices to look up the tables $T$.
The size of $x'$ is determined by the vector dimension in the tables.
Suppose that the vector dimension is $u$ ($u$ is set to 1 in compressed lookup tables); then the size of $x'$ is $m$$\times$$n$$\times$$3u$.
The reason is that each color in each channel of the RGB space is denoted by an integer, but each color in the lookup tables is represented by a vector with $u$ values.
$x'$ is used as the input to the DNN, so that the loss of the Lookup-VNet is:
\begin{equation}
\label{4}
C(f, X, Y, T, W) = L(f(W, X'), Y)
\end{equation}
where $X'$ are the results of $lookup(X,T)$.

Note that in (\ref{4}), $X'$ contain the parameters from the lookup tables $T$.
The weights $W$ and the lookup tables $T$ are learned simultaneously during the whole training process through a gradient descent based optimizer:
\begin{equation}
\label{5}
\Theta_{t+1} = \Theta_{t} - \lambda \frac{\partial L(f(W_{t}, x_{bat}^{'t}), y_{bat}^{t})}{\partial \Theta_{t}}
\end{equation}
where $\Theta = [W, T]$, i.e., the list of all parameters in $W$ and $T$; $\Theta_{t}$ are the values of $\Theta$ in step $t$; $x_{bat}^{'t}$ are the results of $lookup(x_{bat}^{t},T)$; and $\lambda$ is the learning rate.

\subsubsection{Cross-Network Learning}
Beside learning lookup tables on a task with a DNN, lookup tables can also be learned across two or more networks for a task.
Suppose that there are two DNNs $f$ and $g$ with shared lookup tables $T$ for a task with the training data $(X, Y)$.
We alternately optimize the loss functions of $f$ and $g$ on the task with a gradient descent based optimizer, which can be simply written as:
\begin{equation}
\label{6}
\Theta_{t+1}^{f} = \Theta_{t}^{f} - \lambda_f \frac{\partial L(f(W_{t}^f, x_{bat}^{'t}), y_{bat}^{t})}{\partial \Theta_{t}^f}
\end{equation}
where $\Theta^f = [W^f, T]$; $W^f$ and $W^g$  are the weights of $f$ and $g$, respectively; $\Theta_{t}^f$ are the values of $\Theta^f$ in step $t$; and $\lambda_f$ is the learning rate for training $f$. 

\begin{equation}
\label{7}
\Theta_{t+1}^g = \Theta_{t}^{g} - \lambda_g \frac{\partial L(f(W_{t}^g, x_{bat}^{'t}), y_{bat}^{t})}{\partial \Theta_{t}^g}
\end{equation}

By alternately executing (\ref{6}) and (\ref{7}), we learn the lookup tables across two networks $f$ and $g$.

\subsubsection{Cross-Task Learning}
To further explore learning inputs to DNNs, we introduce learning lookup tables across tasks.
Intuitively, the lookup tables learned across two or more tasks are more robust than those learned on one task.
Suppose that $f$ and $g$ are two DNNs with shared tables $T$ for two tasks $p$ and $q$ with the training data $(X_p, Y_p)$ and $(X_q, Y_q)$, respectively.
We alternately optimize the loss functions of $f$ on task $p$ and $g$ on task $q$ with gradient descent, which is written as:
\begin{equation}
\label{8}
\Theta_{t+1}^{f} = \Theta_{t}^{f} - \lambda_f \frac{\partial L(f(W_{t}^f, x_{pbat}^{'t}), y_{pbat}^{t})}{\partial \Theta_{t}^f}
\end{equation}

\begin{equation}
\label{9}
\Theta_{t+1}^g = \Theta_{t}^{g} - \lambda_g \frac{\partial L(f(W_{t}^g, x_{qbat}^{'t}), y_{qbat}^{t})}{\partial \Theta_{t}^g}
\end{equation}

By alternately executing (\ref{8}) and (\ref{9}), the lookup tables are learned across two different tasks $p$ and $q$.

\subsection{Additional Costs of Lookup-VNets Compared with Standard Deep Neural Networks}
In this part, we provide the additional space and computation costs of a Lookup-VNet compared with those of the corresponding standard DNN.

\subsubsection{Additional Space Cost}
When a standard DNN is converted to the corresponding Lookup-VNet with lookup tables of 1-dimension vectors, the whole network architecture remains the same.
The only additional parameters are from three lookup tables with 768 (256$\times$3) parameters.
When the vector dimension is greater than 1, only the first layer of the standard DNN needs to be changed.
Without loss of generality, we assume that the first layer of a standard DNN is a convolutional layer with kernel size $k\times k$ and kernel number $j$. Then the parameter number in the first layer is $k$$\times$$k$$\times$$3$$\times$$j$ where $3$ is the channel number of the input image in the RGB space.
For the corresponding Lookup-VNet, suppose that the vector dimension of the lookup tables is $u$; then the space cost for the three lookup tables is $256$$\times$$3$$\times$$u$.
As each color in each channel is mapped into a vector with dimension $u$, the input channel number is changed from $3$ to $3u$ and the parameter number in the first layer is changed to $k$$\times$$k$$\times$$3u$$\times$$j$.
Therefore, the total additional parameter number is $256$$\times$$3$$\times$$u$+$k$$\times$$k$$\times$$3(u-1)$$\times$$j$.
The experiments suggest that the vector dimension almost has no influence on the performances of Lookup-VNets.
Thus, the additional cost is almost ignorable as we can always take small vector dimensions.
For example, the total parameter number of the standard VGG-16 \cite{Simonyan15} is 138.4 million while the additional parameter number brought by the corresponding Lookup-VNet with vector dimension 1 is only 768 which is ignorable compared with 138.4 million.


\subsubsection{Additional Computation Cost}
The additional computation cost in the forward propagation is related to the input image size.
Suppose that the input image size in the standard DNN is $m$$\times$$n$$\times$$3$ where $m$, $n$, and 3 are the height, the width, and the channel number, respectively, and assume that the computation cost for looking up tables equals to the number of query indices. Then the computation cost for looking up tables is $m$$\times$$n$$\times$$3$.
Suppose both the vertical and horizontal strides in the first convolutional layer are $s$ and the padding strategy is adopted. Then the additional computation cost in the first layer of the Lookup-VNet is $\lceil\frac{m}{s}\rceil\times\lceil\frac{n}{s}\rceil\times j\times(2k^2\times3u+1)-\lceil\frac{m}{s}\rceil\times\lceil\frac{n}{s}\rceil\times j\times(2k^2\times3+1)$ floats.
The total additional cost is $m\times n\times3+\lceil\frac{m}{s}\rceil\times\lceil\frac{n}{s}\rceil\times j\times(2k^2\times3u+1)-\lceil\frac{m}{s}\rceil\times\lceil\frac{n}{s}\rceil\times j\times(2k^2\times3+1)$ floats.
Note that when the vector dimension $u$ is set to 1, the only additional cost is $m\times n\times3$ for looking up tables.

\section{Experiments}
\label{exp}

In the section, we report a series of experiments conducted on image classification tasks to explore deep collective learning in computer vision with Lookup-VNets.
The code will be released.
Through these experiments, we intend to answer the following questions:
\begin{itemize}
\item \textbf{Q1:} How much is the accuracy gap between Lookup-VNets and the corresponding standard DNNs?
\item \textbf{Q2:} How does the vector dimension in the lookup tables influence the performances of Lookup-VNets?
\item \textbf{Q3:} How does the CMP-Rate (the rate of compressing the color space) influence the performances of Lookup-VNets?
\item \textbf{Q4:} How does the table learning strategy influence the the performances of Lookup-VNets?
\item \textbf{Q5:} How do images look like when they are represented by the learned lookup tables?
\end{itemize}

\begin{table*}[]
\caption{Test Accuracies (\%) on CIFAR-10 and CIFAR-100 with Full Lookup Tables of Different Vector Dimensions}
\begin{center}
\label{m1}
\begin{tabular}{cccccccccc}
\hline
                           &           & Standard & Dim 1 & Dim 2 & Dim 3 & Dim 4 & Dim 5 & Dim 10 & Dim 100 \\ \hline
\multirow{3}{*}{CIFAR-10}  & ResNet-20 & 91.28    & 91.52 & 91.72 & 91.17 & 91.29 & 91.23 & 91.01  & 91.33   \\
                           & VGG-16    & 93.20    & 93.29 & 93.27 & 93.13 & 93.10 & 93.38 & 93.12  & 93.28   \\
                           & WRN-40-4  & 95.10    & 95.26 & 94.83 & 95.11 & 95.10 & 94.95 & 95.01  & 94.95   \\ \hline
\multirow{3}{*}{CIFAR-100} & ResNet-20 & 66.63    & 66.37 & 66.51 & 66.29 & 66.65 & 66.14 & 66.19  & 66.67   \\
                           & VGG-16    & 72.48    & 72.17 & 72.16 & 72.46 & 72.42 & 72.20 & 72.52  & 72,32   \\
                           & WRN-40-4  & 77.97    & 77.60 & 78.14 & 78.06 & 77.50 & 77.82 & 77.48  & 77.34   \\ \hline
\end{tabular}
\end{center}
\end{table*}

\begin{table*}[!t]
\caption{Test Accuracies (\%) on Tiny ImageNet with Full Lookup Tables of Different Vector Dimensions}
\centering
\label{m3}
\begin{tabular}{cccccccccccc}
\hline
          & Standard & Dim 1 & Dim 5 & Dim 10 & Dim 20 & Dim 30 & Dim 50 & Dim 100 \\ \hline
ResNet-20 & 50.69    & 50.68  & 50.31 & 50.05  & 50.21  & 50.55  & 50.36  & 50.21   \\ \hline
VGG-16    & 60.61    & 61.15  & 60.50 & 61.13  & 61.10  & 60.18  & 61.47  & 61.50   \\ \hline
WRN-40-4  & 63.04    & 63.02  & 62.67 & 63.11  & 63.22  & 63.11   & 62.64  & 63.03   \\ \hline
\end{tabular}
\end{table*}

\subsection{Datasets}
In the experiments, we adopt four benchmark datasets: CIFAR-10 \cite{krizhevsky2009learning}, CIFAR-100 
\cite{krizhevsky2009learning}, Tiny ImageNet \footnote{http://tiny-imagenet.herokuapp.com/}, and ImageNet (ILSVRC2012) \cite{deng2009imagenet}.

$\textbf{CIFAR-10}$ is an image classification dataset with 10 classes, containing 50,000 training images and 10,000 test images with image size 32 $\times$ 32 in the RGB space.
We follow the standard data augmentation on CIFAR datasets.
During training time, we pad 4 pixels on each side of an image and randomly flip it horizontally.
Then the image is randomly cropped to 32 $\times$ 32 size. 
During test time, we only evaluate the single view of an original 32 $\times$ 32 image without padding or cropping.

$\textbf{CIFAR-100}$ comprises similar images to those in CIFAR-10, but has 100 classes.
We adopt the same data augmentation strategy as that in CIFAR-10.

$\textbf{Tiny ImageNet}$, i.e., a subeset of ImageNet, is an image classification dataset with 200 classes, containing 100,000 training images and 10,000 test images with size 64 $\times$ 64 in the RGB space.
At training time, we pad 8 pixels on each side of an image and randomly flip it horizontally, then the image is randomly cropped to 64 $\times$ 64 size.
At test time, we only evaluate the original image.

$\textbf{ImageNet}$ is a large-scale image classification dataset with 1000 classes, containing 1.28 million training images and 50,000 validation images with different sizes in the RGB space.
On ImageNet, to reduce the CPU burden, we adopt a simpler data augmentation strategy than that in the models pretrained by Facebook \footnote{https://github.com/facebookarchive/fb.resnet.torch}.
Specifically, we use a simple scale and aspect ratio augmentation strategy from \cite{szegedy2015going}.
Test images are resized so that the shorter side is set to 256, and then are cropped to size 224 $\times$ 224.\par

Note that in Lookup-VNets, data preprocessing is not needed as the inputs are learned end-to-end .
However, in standard DNNs, data preprocessing is necessary as their inputs are images which are represented by 8-bit integers ranging from 0 to 255.
Therefore, to ensure the performances of standard DNNs (the baselines), we use data preprocessing for them.
On CIFAR and Tiny-ImageNet datasets, we use the widely used data proprocessing strategy: each image is preprocessed by subtracting its mean and dividing it by its standard deviation.
On ImageNet, each image is preprocessed by subtracting the mean of the whole training set and dividing it by the standard deviation.\par

In every case below, the experiments are repeated three times and then we report the average test accuracy as the variance is quite small.

\begin{table}[!t]
\caption{Validation Accuracies (\%) on ImageNet of Standard DNNs and Lookup-VNets with 1-Dimension Full Lookup Tables}
\centering
\label{m4}
\begin{tabular}{cccc}
\hline
                                                                      &          & \#Parameters & Accuracy \\ \hline
\multirow{2}{*}{\begin{tabular}[c]{@{}c@{}}Resnet-18\end{tabular}} & Standard & 11.7 M       & 69.4 \\
                                                                      & Dim 1    & 11.7 M   &  \textbf{70.1}          \\ \hline
\multirow{2}{*}{\begin{tabular}[c]{@{}c@{}}Resnet-34\end{tabular}} & Standard & 21.8 M       & 73.0  \\
                                                                      & Dim 1    & 21.8 M       &\textbf{73.4}           \\ \hline
\end{tabular}
\end{table}

\begin{figure*}[!t]
   \begin{minipage}{0.32\textwidth}
     \centering
         \includegraphics[height=4.13cm, width=5.5cm]{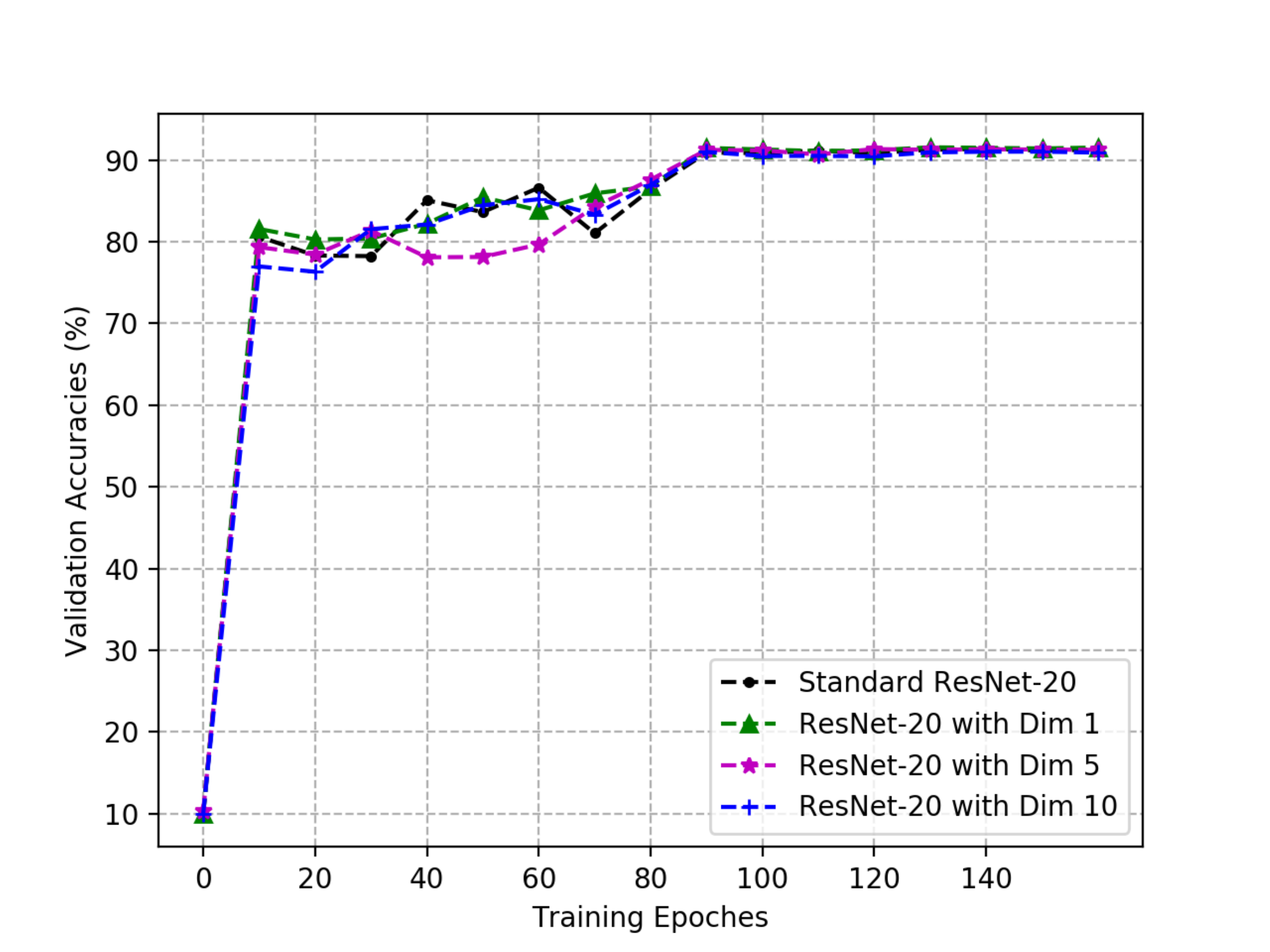}
     \caption{Training Curves of ResNet-20 on CIFAR-10 with Different Vector Dimensions}
     \label{f4_1}
   \end{minipage}\hfill
   \begin{minipage}{0.32\textwidth}
     \centering
     \includegraphics[height=4.13cm, width=5.5cm]{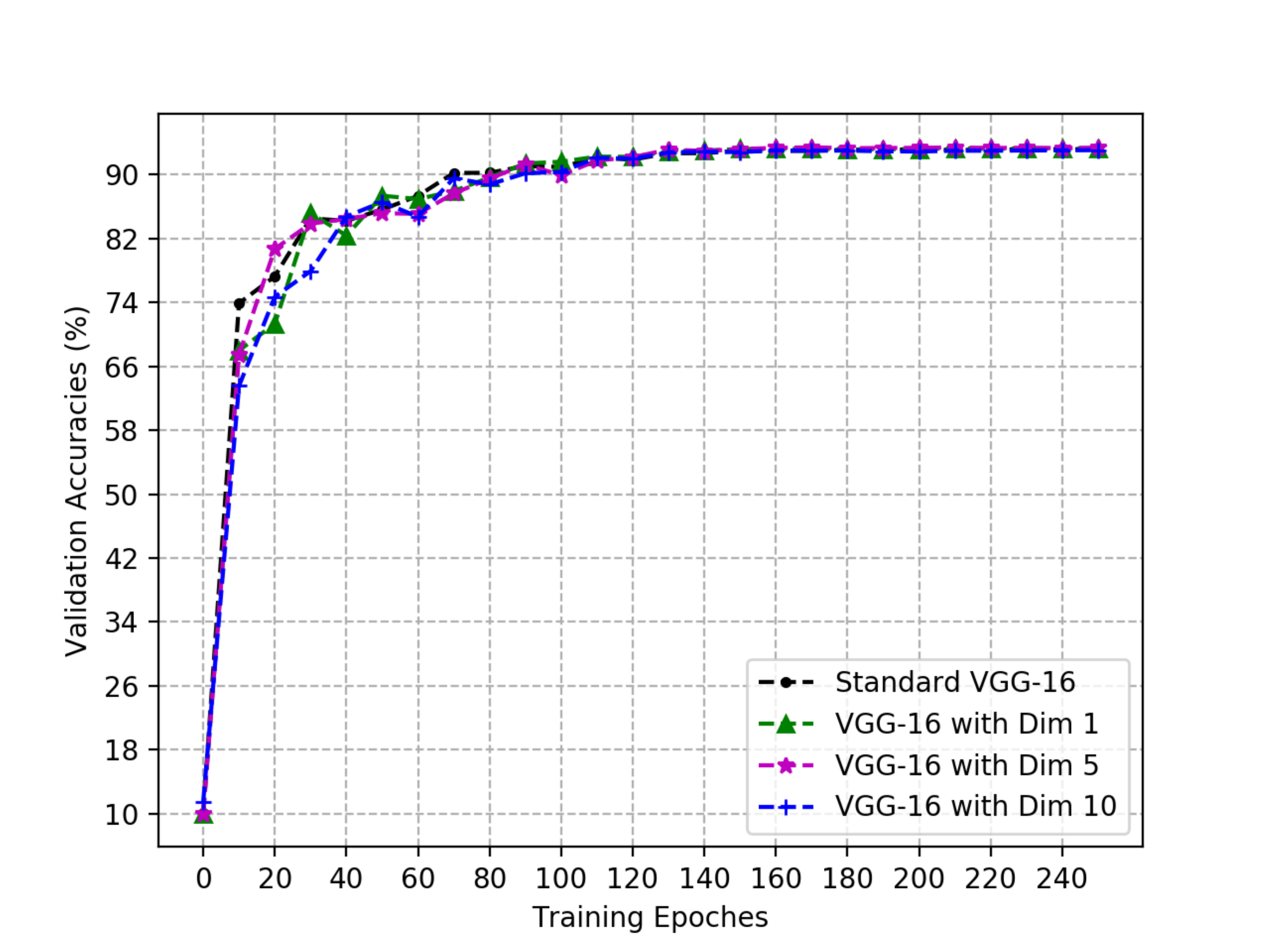}
     \caption{Training Curves of VGG-16 on CIFAR-10 with Different Vector Dimensions}
     \label{f4_2}
   \end{minipage}\hfill
   \begin{minipage}{0.32\textwidth}
     \centering
          \includegraphics[height=4.13cm, width=5.5cm]{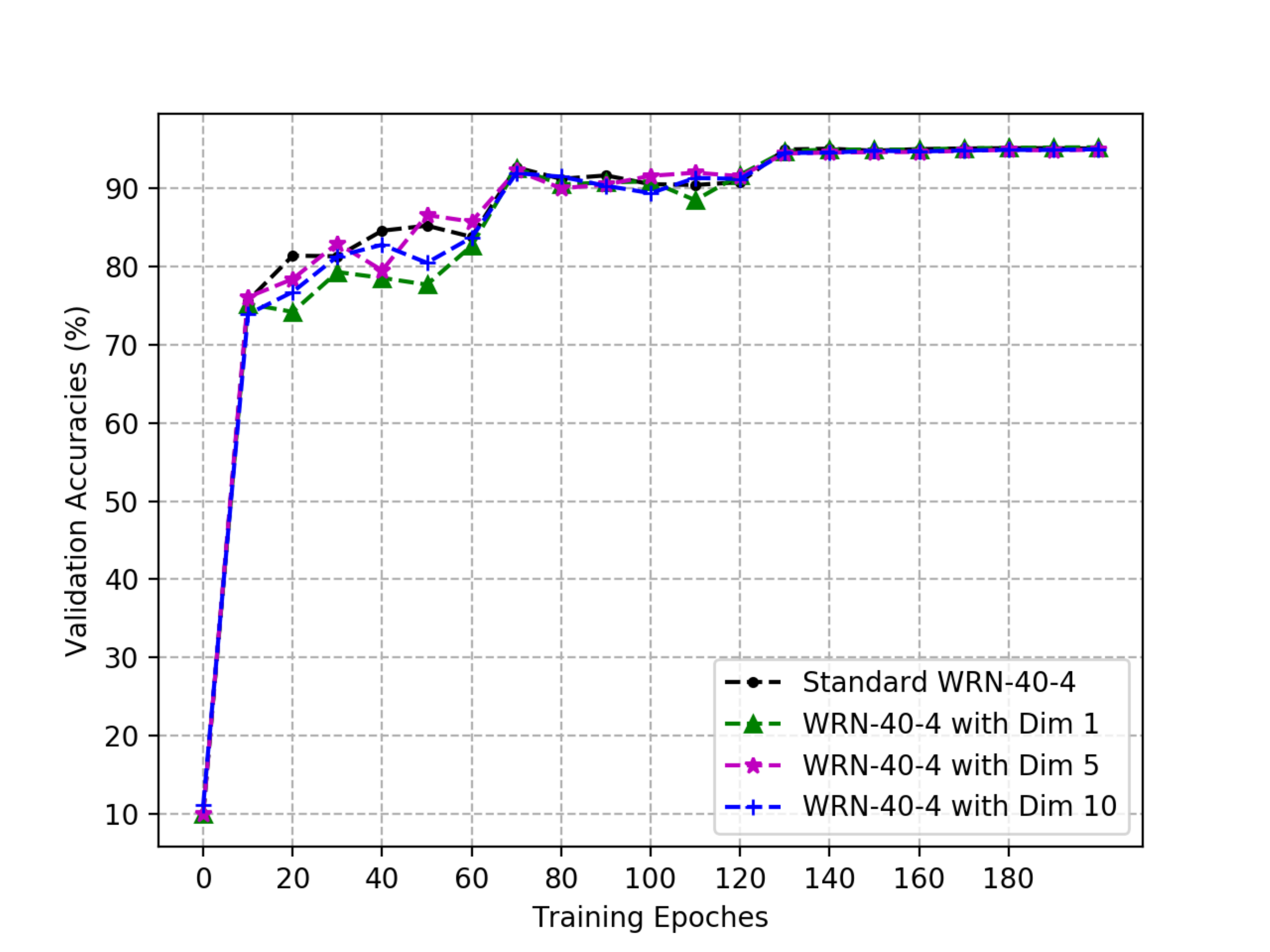}
     \caption{Training Curves of WRN-40-4 on CIFAR-10 with Different Vector Dimensions}
     \label{f4_3}     
   \end{minipage}  
\end{figure*}

\begin{figure*}[]
  \vskip 0.2in
   \begin{minipage}{0.32\textwidth}
     \centering
     \includegraphics[height=4.13cm, width=5.5cm]{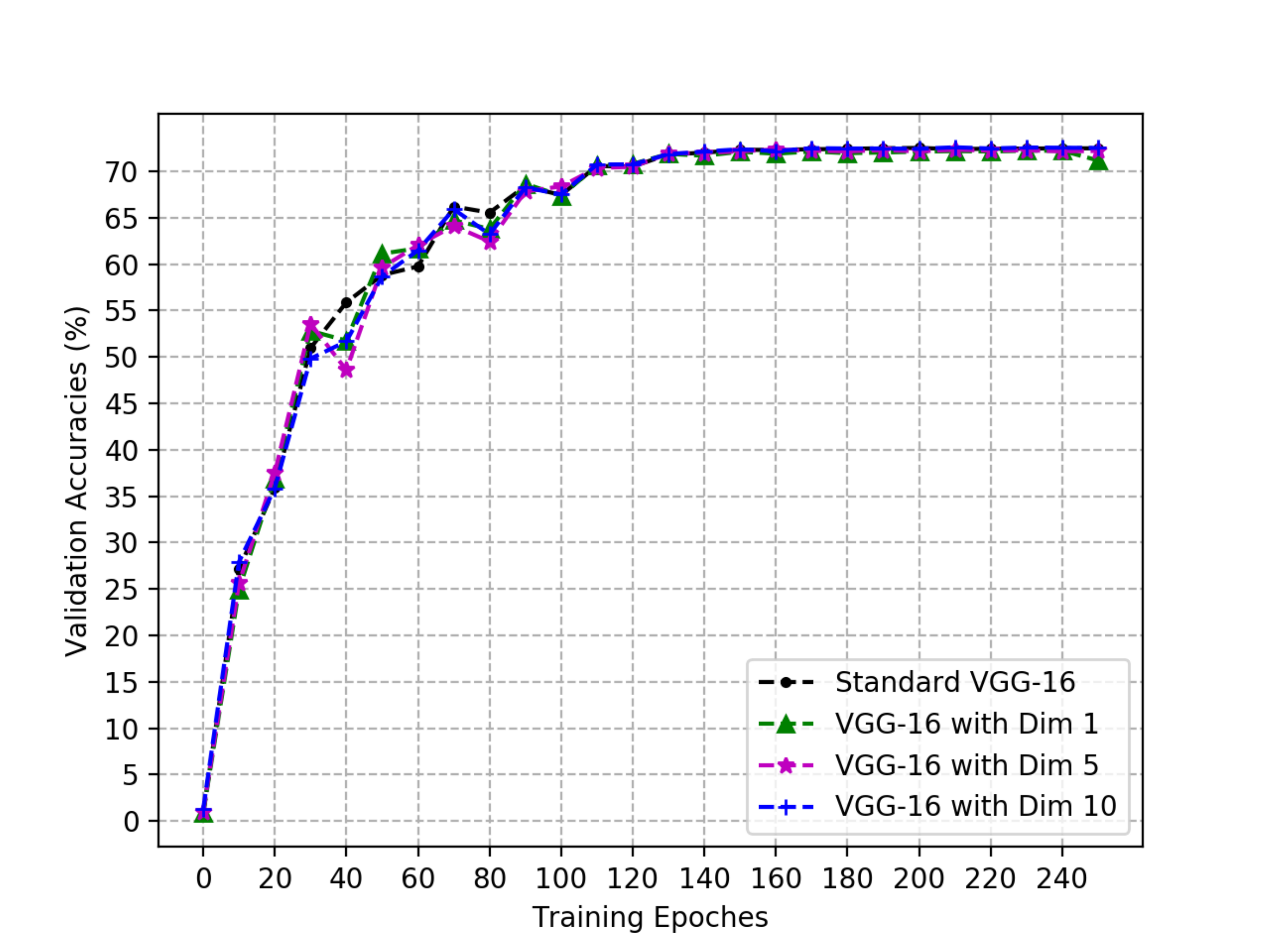}
     \caption{Training Curves of VGG-16 on CIFAR-100 with Different Vector Dimensions}\label{f44_1}
   \end{minipage}\hfill
   \begin{minipage}{0.32\textwidth}
     \centering
     \includegraphics[height=4.13cm, width=5.5cm]{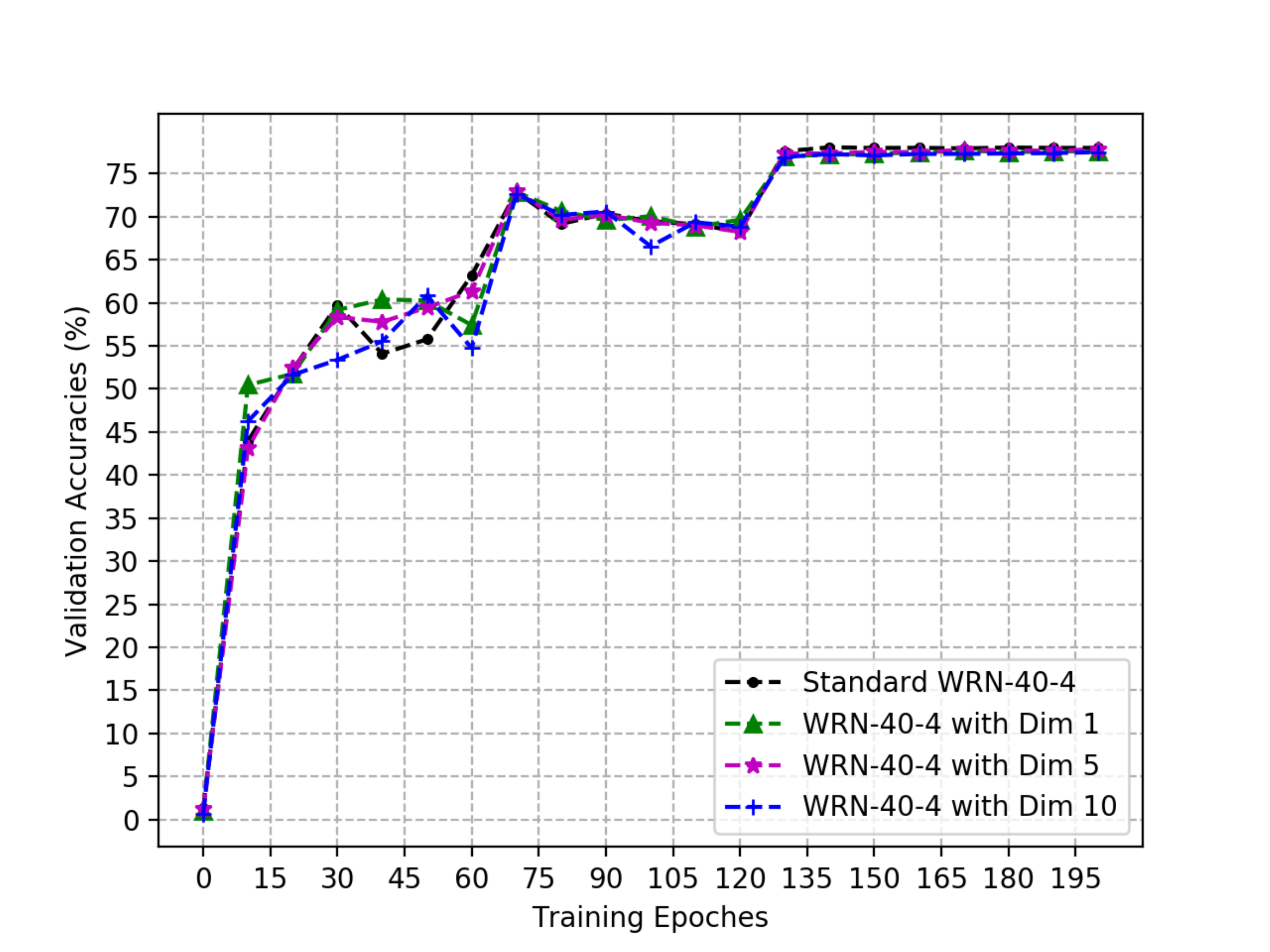}
     \caption{Training Curves of WRN-40-4 on CIFAR-100 with Different Vector Dimensions}\label{f44_2}
   \end{minipage}\hfill
   \begin{minipage}{0.32\textwidth}
     \centering
     \includegraphics[height=4.13cm, width=5.5cm]{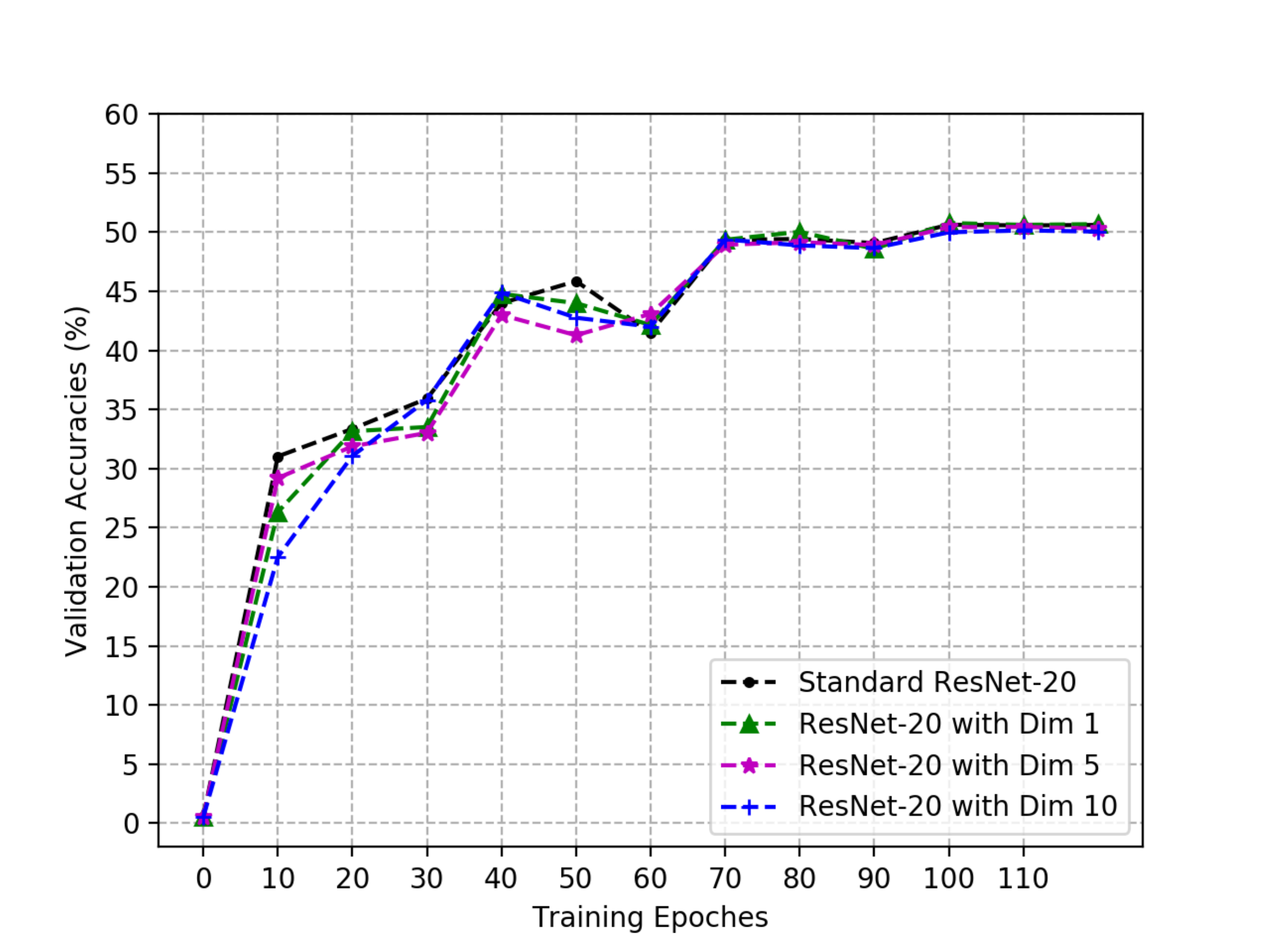}
     \caption{Training Curves of ResNet-20 on Tiny ImageNet with Different Vector Dimensions}
     \label{f5_1}
   \end{minipage}   
   \vskip -0.2in
\end{figure*}

\begin{figure*}[]
  \vskip 0.2in
     \begin{minipage}{0.32\textwidth}
     \centering
     \includegraphics[height=4.13cm, width=5.5cm]{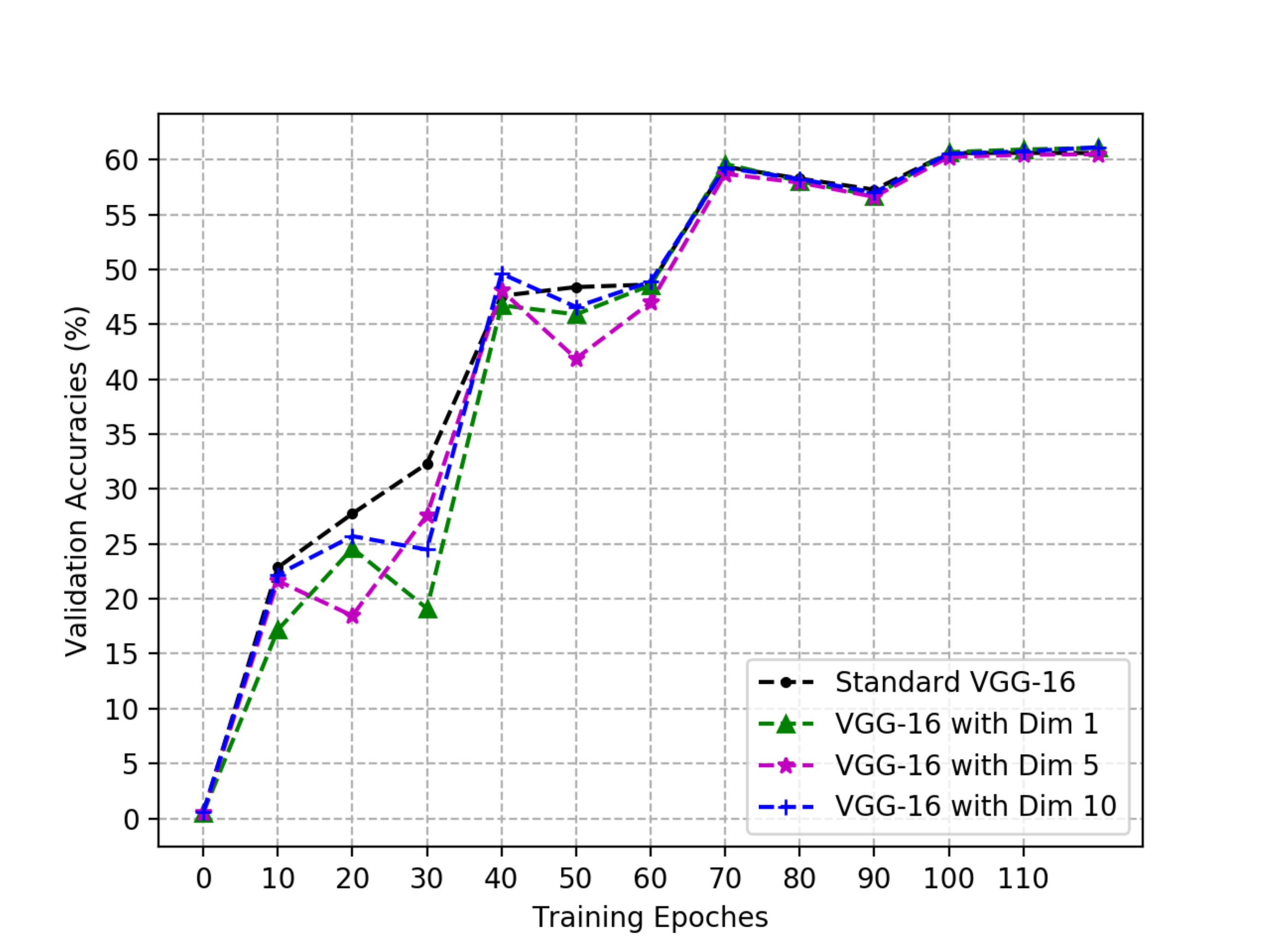}
     \caption{Training Curves of VGG-16 on Tiny ImageNet with Different Vector Dimensions}
     \label{f5_2}
   \end{minipage}\hfill
   \begin{minipage}{0.32\textwidth}
     \centering
     \includegraphics[height=4.13cm, width=5.5cm]{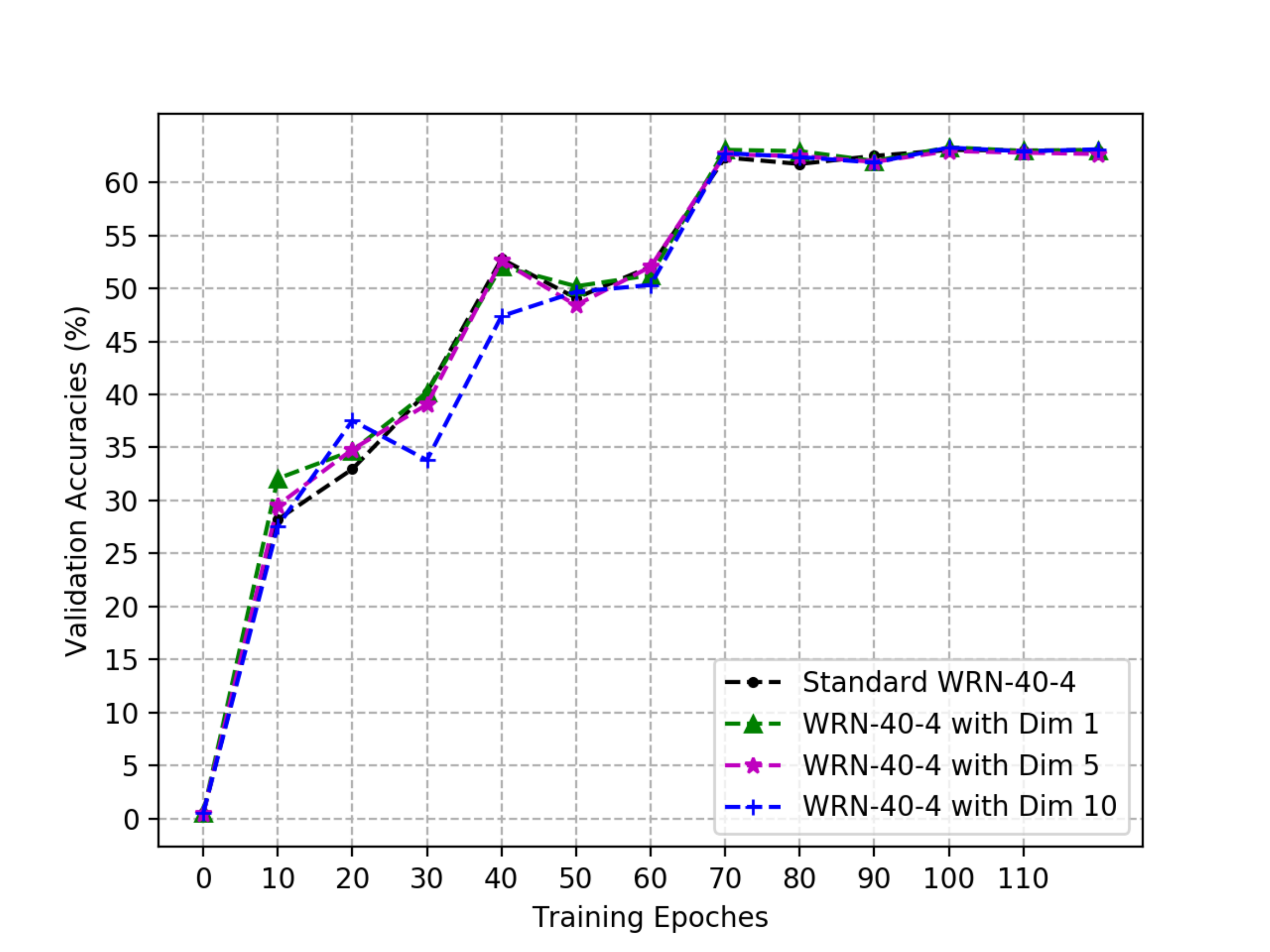}
     \caption{Training Curves of WRN-40-4 on Tiny ImageNet with Different Vector Dimensions}
      \label{f5_3}
   \end{minipage}\hfill
   \begin{minipage}{0.32\textwidth}
     \centering
     \includegraphics[height=4.13cm, width=5.5cm]{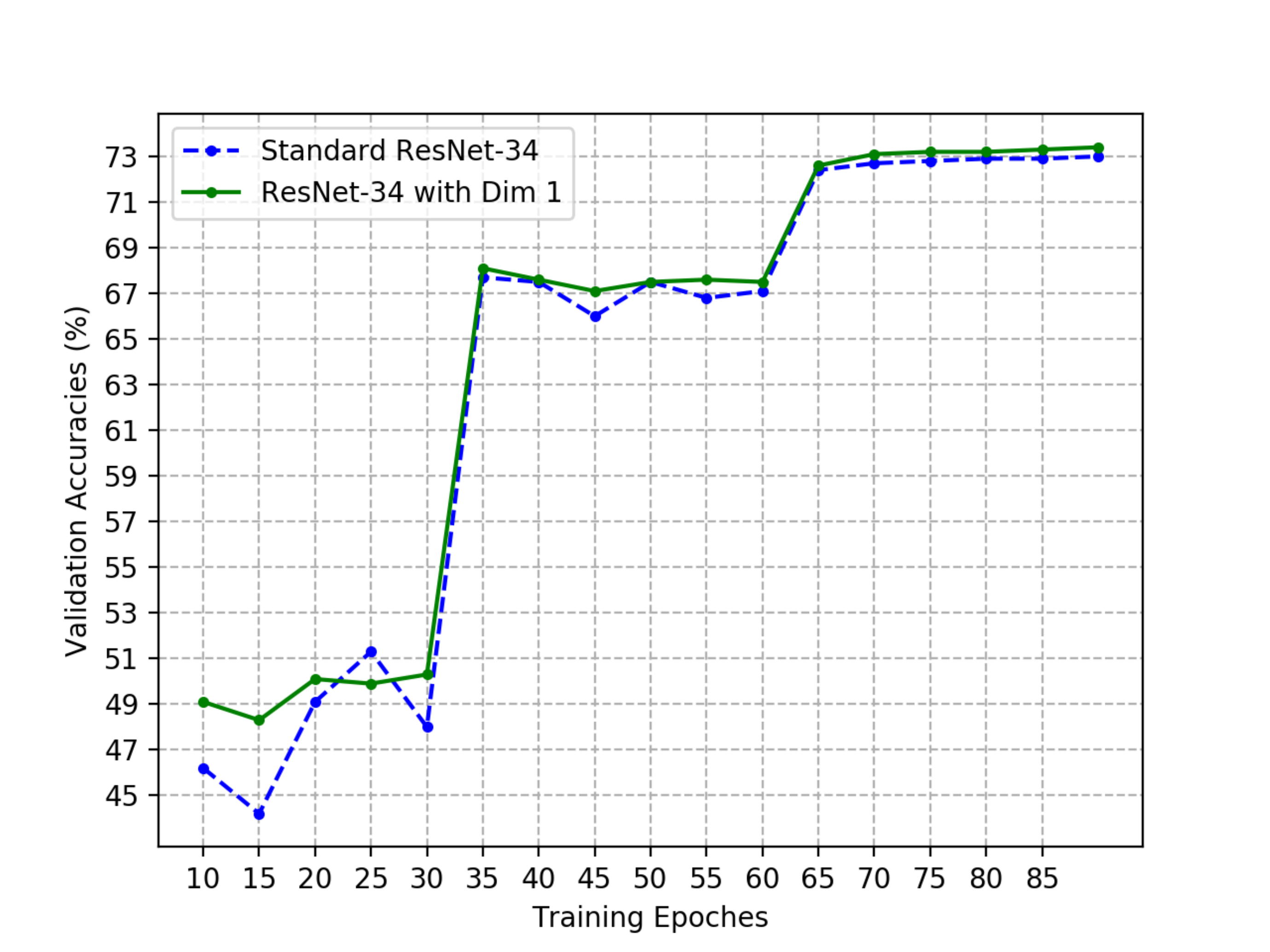}
     \caption{Training Curves of ResNet-34 on ImageNet with Vector Dimension 1}
     \label{f6_1}
   \end{minipage}
   \vskip -0.2in
\end{figure*}

\subsection{Architectures and Experiment Setup}
We adopt the widely used network architectures ResNet \cite{he2016deep}, VGG \cite{Simonyan15}, and WRN \cite{Zagoruyko2016WRN}.
Specifically, ResNet-20, VGG-16, and WRN-40-4 are adopted for CIFAR-10, CIFAR-100, and Tiny ImageNet, and ResNet-18 and ResNet-34 are used for ImageNet.
VGG-16 is modified for small images as suggested by \cite{liu2015very}, i.e., replacing the two fully connected layers of 4096 neurons with one fully connected layer with 512 neurons.\par

On CIFAR-10 and CIFAR-100, we use the training hyperparameters in the original studies to train ResNet-20 and WRN-40-4.  
For VGG-16, we have trained it for 250 epochs with mini-batch size 128 and optimizer SGD with moment 0.9;
the weight decay is set to 5e-4;
the initial learning rate is set to 0.1 and divided by 2 after every 20 epochs.
\par
On Tiny ImageNet, the hyperparameters in weight decay for ResNet-20, VGG-16, and WRN-40-4 are set to 1e-4, 5e-4, and 5e-4, respectively.
We have trained the three models for 120 epochs with mini-batch size 128 and optimizer SGD with moment 0.9.
The initial learning rate is 0.05 and divided by 5 after every 30 epochs.\par
On the large-scale dataset ImageNet, we have trained ResNet-18 and ResNet-34 for 90 epochs with optimizer SGD with moment 0.9.
The mini-batch size is set to 128 and the learning rate is set to 0.05 and divided by 10 after every 30 epochs for both networks.

\begin{figure*}[]
   \begin{minipage}{0.32\textwidth}
     \centering
     \includegraphics[height=4.13cm, width=5.5cm]{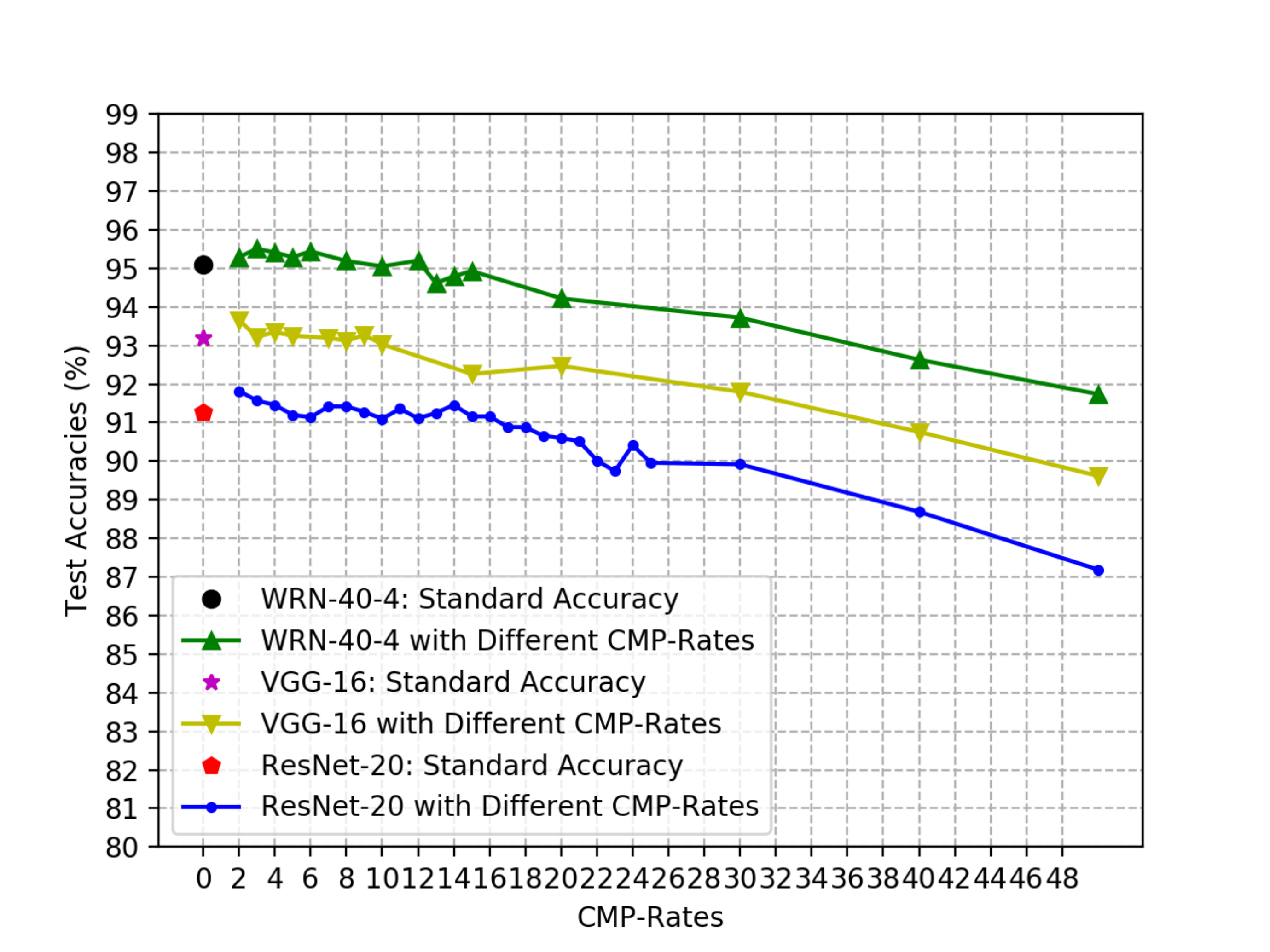}
     \caption{Test Accuracies (\%) on CIFAR-10 with Different CMP-Rates}
     \label{f7}
   \end{minipage}\hfill
   \begin{minipage}{0.32\textwidth}
     \centering
     \includegraphics[height=4.13cm, width=5.5cm]{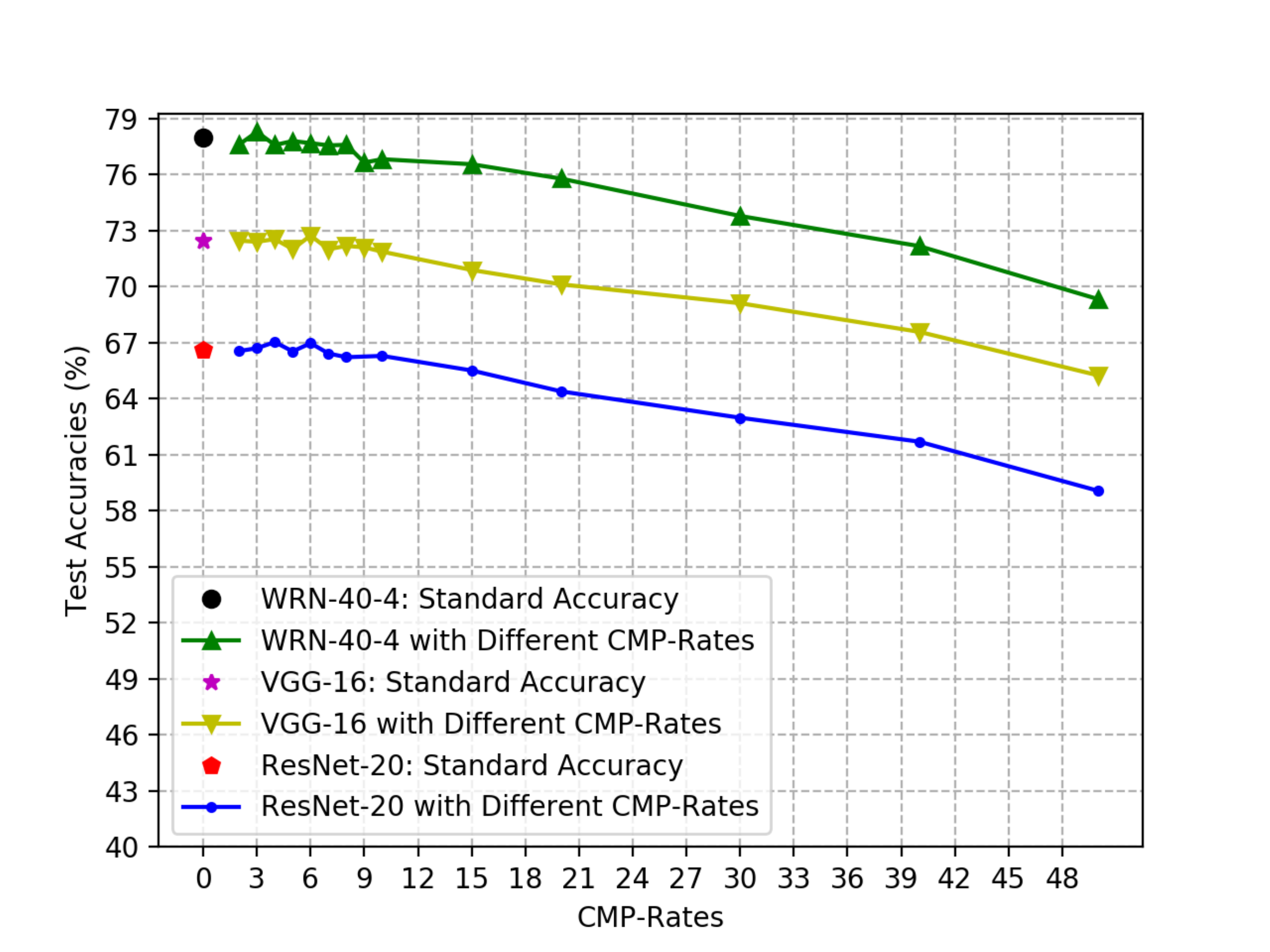}
     \caption{Test Accuracies (\%) on CIFAR-100 with Different CMP-Rates}
    \label{f8}
   \end{minipage}\hfill
    \begin{minipage}{0.32\textwidth}
     \centering
     \includegraphics[height=4.13cm, width=5.5cm]{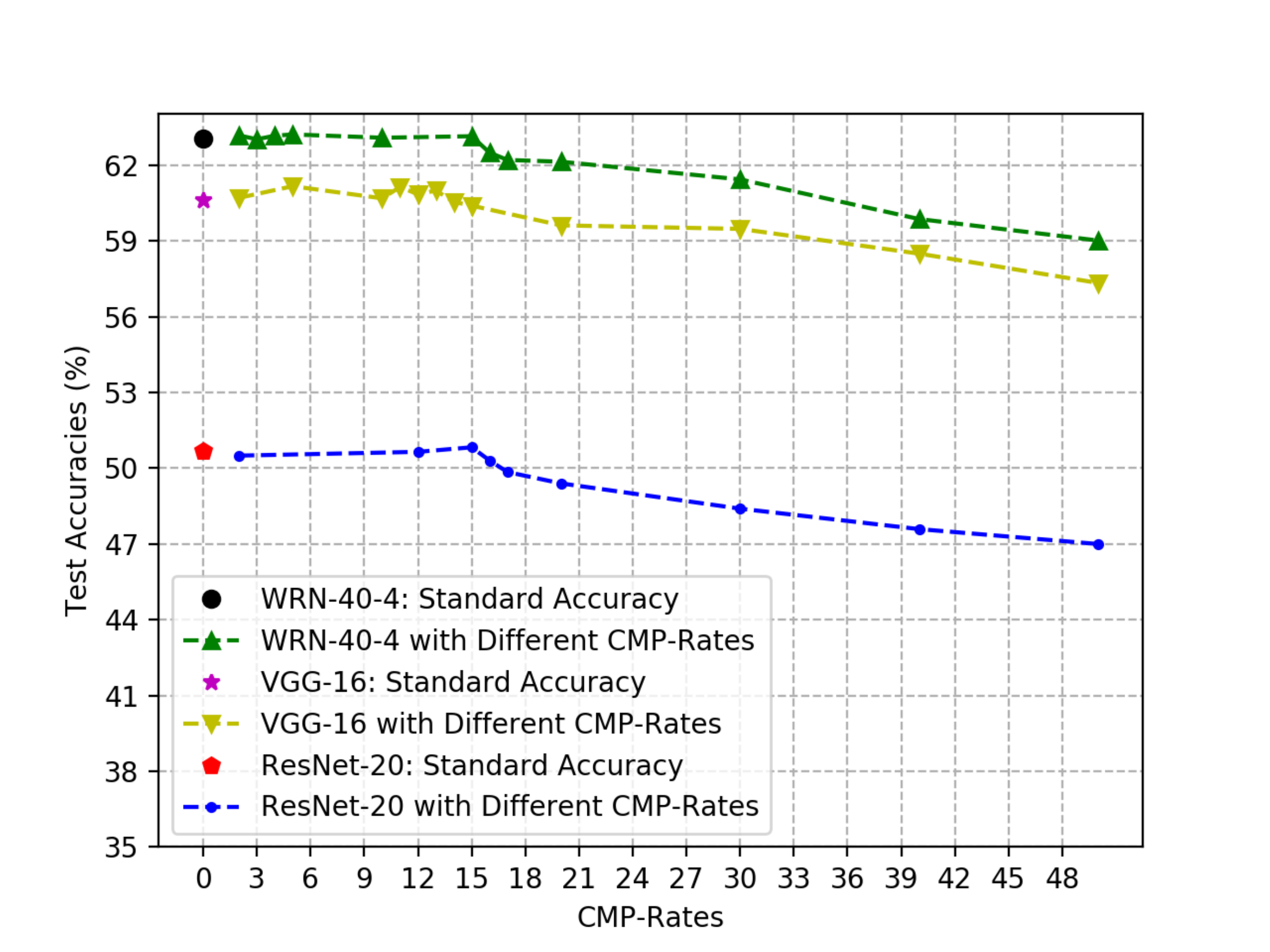}
     \caption{Test Accuracies (\%) on Tiny ImageNet with Different CMP-Rates}
    \label{f9}
   \end{minipage}  
\end{figure*}

\begin{table*}[]
\caption{Test Accuracies (\%) on CIFAR-10, CIFAR-100, and Tiny ImageNet of Lookup-VNets with Large CMP-Rates}
\label{m5}
\centering
\begin{tabular}{cccccc}
\hline
                               &           & Random Guess & CMP-Rate 100 & CMP-Rate 128 & CMP-Rate 256 \\ \hline
\multirow{3}{*}{CIFAR-10}      & ResNet-20 & 10.00           & 80.77        & 75.30        & 10.00        \\
                               & VGG-16    & 10.00           & 82.66        & 78.94        & 10.00        \\
                               & WRN-40-4  & 10.00           & 85.85        & 80.49        & 10.00        \\ \hline
\multirow{3}{*}{CIFAR-100}     & ResNet-20 & 1.00            & 50.96        & 43.69        & 1.00         \\
                               & VGG-16    & 1.00            & 55.31        & 47.54        & 1.00         \\
                               & WRN-40-4  & 1.00            & 59.85        & 51.05        & 1.00         \\ \hline
\multirow{3}{*}{Tiny ImageNet} & ResNet-20 & 0.50          & 40.41        & 34.37        & 0.50         \\
                               & VGG-16    & 0.50          & 49.53        & 41.78        & 0.50         \\
                               & WRN-40-4  & 0.50          & 49.40        & 41.20        & 0.50         \\ \hline
\end{tabular}
\end{table*}

\begin{table}[]
\caption{Validation Accuracies (\%) on ImageNet with Different CMP-Rates}
\label{m5_2}
\centering
\begin{tabular}{cccc}
\hline
                           &             & \#Parameter & Accuracy \\ \hline
\multirow{4}{*}{ResNet-18} & Standard    & 11.7 M     & 69.4     \\
                           & CMP-Rate 2  & 11.7 M     & 69.9         \\
                           & CMP-Rate 10 & 11.7 M    & 69.7         \\
                           & CMP-Rate 15 & 11.7 M     &69.4          \\ \hline
\multirow{4}{*}{ResNet-34} & Standard    & 21.8 M      & 73.0     \\
                           & CMP-Rate 2  & 21.8 M      & 73.2     \\
                           & CMP-Rate 10  & 21.8 M     & 73.1        \\
                           & CMP-Rate 15 & 21.8 M      &73.0          \\ \hline
\end{tabular}
\end{table}

\begin{table*}[!t]
\caption{Test Accuracies (\%) on CIFAR-10(100) and Tiny ImageNet with Full Lookup Tables Learned across ResNet-20 and VGG-16}
\centering
\label{m6}
\begin{tabular}{ccc|cc|cc}
\hline
         & \multicolumn{2}{c|}{CIFAR-10} & \multicolumn{2}{c|}{CIFAR-100} & \multicolumn{2}{c}{Tiny ImageNet} \\ \hline
         & ResNet-20       & VGG-16      & ResNet-20       & VGG-16       & ResNet-20         & VGG-16        \\ \hline
Standard &   91.28       & 93.20        &66.63         &72.48            & 50.69              &60.61               \\ \hline
Dim 1    &  91.47        &93.48         &66.70         &72.60            & 51.00              & 60.83            \\ \hline
Dim 5    & 91.10        &93.21            & 66.73       &72.52              &50.68              & 61.09         \\ \hline
Dim 10   & 91.01        &93.20            & 66.67         &72.25             & 50.55             &61.49               \\ \hline
Dim 100  & 91.28        & 93.20            & 66.87         &72.53              &50.64           & 60.50              \\ \hline
\end{tabular}
\end{table*}
\subsection{Performances of Lookup-VNets with Full Lookup Tables of Different Vector Dimensions}
Lookup-VNets parameterize the inputs through associating each color with a vector in lookup tables.
To investigate the influence of the vector dimension on the performances of Lookup-VNets, substantial experiments with various vector dimensions are conducted on the four datasets.
Specifically, we evaluate Lookup-VNets with different vector dimensions on CIFAR-10, CIFAR-100, and Tiny ImageNet while we only check the performances of Lookup-VNets with vector dimension 1 on ImageNet due to the large image size.
Full lookup tables are initialized evenly between -1 and 1. \par

Table \ref{m1} and Table \ref{m3} summarize the performances of the standard DNNs and the corresponding Lookup-VNets with different vector dimensions on CIFAR (i.e, CIFAR-10 and CIFAR-100) and Tiny ImageNet, respectively.
Surprisingly, Lookup-VNets with different vector dimensions have almost the same performance with a given network architecture.
This indicates that the vector dimension has almost no influence on the performances (i.e., generalization ability \footnote{As their training accuracies are all 100\%, then test performances represent the generalization abilities (Generalization = Training Accuracy - Test Accuracy).} ) of Lookup-VNets while the network architecture matters.
Another astonishing observation is that these Lookup-VNets produce almost the same results as those of the corresponding standard DNNs with RGB inputs on the three datasets.
We attributed this observation to the small number of training data in the three datasets because a different phenomenon is observed on the large-scale and challenging dataset ImageNet as shown in Table \ref{m4}.


As seen from Table \ref{m4}, 1-dimension Lookup-VNets show a consistent performance improvement on ImageNet with ResNet-18 and ResNet-34, which indicates that Lookup-VNets have advantages over the standard DNNs on large-scale and challenging datasets like ImageNet.
The possible reason can be that when the the size and complexity of the dataset are scaled up, the integers in the RGB space are not appropriate for DNN training anymore, but the Lookup-VNets make the inputs to DNNs more flexible, and are able to learn the optimal inputs automatically.
It is worth noting that the number of the additional parameters brought by 1-dimension Lookup-VNets is only 768 which is ignorable compared with 11.7 million parameters in ResNet-18 and 21.8 million parameters in ResNet-34.
\par

Figure \ref{f4_1}, Figure \ref{f4_2}, and  Figure \ref{f4_3}  present the training curves of ResNet-20, VGG-16, and WRN-40-4 with different vector dimensions on CIFAR-10, respectively.
Figure \ref{f44_1} and Figure \ref{f44_2} present the training curves of VGG-16 and WRN-40-4 with different vector dimensions on CIFAR-100, respectively. 
Figure \ref{f5_1}, Figure \ref{f5_2} and Figure \ref{f5_3} present the training curves of ResNet-20, VGG-16, and WRN-40-4 with different vector dimensions Tiny ImageNet, respectively.
Figure \ref{f6_1} presents the training curve of ResNet-34 with vector dimension 1 on ImageNet.
It is observed that the Lookup-VNets with different vector dimensions have the same converge speeds as those of the standard DNNs on all the four datasets.

\subsection{Performances of Lookup-VNets with Different CMP-Rates}
Empirical results have shown that the vector dimension in full lookup tables plays no role in the performances of Lookup-VNets, which questions the necessity of a large color space.
Now we explore whether compressing the color space influences the performances of Lookup-VNets.
We compare the performances of the standard DNNs with those of the corresponding Lookup-VNets with various CMP-Rates on CIFAR-10, CIFAR-100, and Tiny ImageNet.
Compressed lookup tables are initialized evenly between -1 and 1. 

\par

Figure \ref{f7}, Figure \ref{f8}, and Figure \ref{f9} represent the results with different CMP-Rates on CIFAR-10, CIFAR-100, and Tiny ImageNet, respectively.
We notice that the performances do not drop compared with those of the standard DNNs with RGB inputs when the compressing rate is less than a threshold.
To our surprise, without accuracy dropping, the color space can be compressed 4096 ($16^3$), 729 ($9^3$), and 1728 ($12^3$) times on CIFAR-10 with ResNet-20, VGG-16, and WRN-40-4, respectively, 512 ($8^3$) times on CIFAR-100 with the three networks, and 3375 ($15^3$), 2744 ($14^3$), and 3375 ($15^3$) times on Tiny ImageNet with ResNet-20, VGG-16, and WRN-40-4, respectively.
The experimental results indicate that Lookup-VNets can be used to learn the optimal image coding scheme for given tasks and goals such as the storage compression and the accuracy. For example, when the color space can be compressed 4096 ($16^3$) times without accuracy dropping, the pixel bits can be reduced from 24 (8 $\times$ 3) bits to 12 ( 4 $\times$ 3) bits under this setting so that the image storage space can be saved one half.

\par


\begin{table}[!t]
\caption{Test Accuracies (\%) of ResNet-20 and VGG-16 with Full Lookup Tables Learned across CIFAR-10 and Tiny ImageNet}
\centering
\label{m7}
\begin{tabular}{ccc|cc}
\hline
         & \multicolumn{2}{c|}{ResNet-20}                                                                                & \multicolumn{2}{c}{VGG-16}                                                                                    \\ \hline
         & \begin{tabular}[c]{@{}c@{}}CIFAR\\ -10\end{tabular} & \begin{tabular}[c]{@{}c@{}}Tiny\\ ImageNet\end{tabular} & \begin{tabular}[c]{@{}c@{}}CIFAR\\ -10\end{tabular} & \begin{tabular}[c]{@{}c@{}}Tiny\\ ImageNet\end{tabular} \\ \hline
Standard & 91.28                                               & 50.69                                                   & 93.20                                               & 60.61                                                   \\
Dim 1    & \textbf{91.93}                                      & 50.69                                                   & 93.21                                               & 61.10                                                   \\
Dim 5    & 91.50                                               & 50.59                                                   & 93.20                                               & \textbf{61.77}                                          \\
Dim 10   & 91.73                                               & 50.61                                                   & 93.21                                               & 60.59                                                   \\ \hline
\end{tabular}
\end{table}

To further explore color space compression, we report several experiments with large CMP-Rates ($\geq100$), i.e., 100, 128, and 256.
The results are presented in Table \ref{m5}.
As expected, the accuracies of Lookup-VNets with CMP-Rate 256, i.e., 1 color in each channel, are as the same as those of the random guess since there is only one color for all images.
However, to our surprise, when the color number in each channel is only increased from 1 (i.e., CMP-Rate 256) to 2 and 3 (i.e., CMP-Rate 128 and 100), the performances are improved substantially on the three datasets, i.e., from 10\% to 75.30\% and 80.77\%, from 10\% to 78.94\% and 82.66\%, and from 10\% to 80.49\% and 85.85\% on CIFAR-10 with ResNet-20, VGG-16, and WRN-40-4, respectively;
from 1\% to 43.69\% and 50.96\%, from 1\% to 47.54\% and 55.31\%, and from 1\% to 51.05\% and 59.85\% on CIFAR-100 with ResNet-20, VGG-16, and WRN-40-4, respectively; from 0.5\% to 34.37\% and 40.41\%, from 0.5\% to 41.78\% and 49.53\%, and from 0.5\% to 41.20\% and 49.40\% on Tiny ImageNet with ResNet-20, VGG-16, and WRN-40-4, respectively.
\par

We also report the experimental results on large-large dataset ImageNet with different CMP-Rates.
As shown in Table \ref{m5_2}, when the color space is compressed 8 ($2^3$), 1000 ($10^3$), and 3375 ($15^3$) times, Lookup-VNets are still able to perform no worse than the standard DNNs, which indicates the promise of Lookup-VNets and the potential of deep collective learning in computer vision.

\subsection{Performances of Lookup-VNets with Different Table Learning Strategies}
In this part, we investigate how the table learning strategy influences the performances of Lookup-VNets.
We consider the classification tasks on CIFAR-10, CIFAR-100, and Tiny ImageNet as three distinct tasks.
For cross-network learning, we learn the lookup tables across ResNet-20 and VGG-16 for each of the three tasks.
For cross-task learning, we use ResNet-20 and VGG-16 on CIFAR-10 and Tiny ImageNet to jointly learn the lookup tables.
Table \ref{m6} and Table \ref{m7} report the results of the lookup tables learned across two networks and across two tasks, respectively.
Compared with the results of the individually learned tables as shown in Table \ref{m1} and Table \ref{m3}, we observe that learning across networks has almost no influence on the performances of Lookup-VNets while learning across tasks is able to improve the performances.

\begin{figure*}[!t]
\centerline{\includegraphics[totalheight=5.8cm]{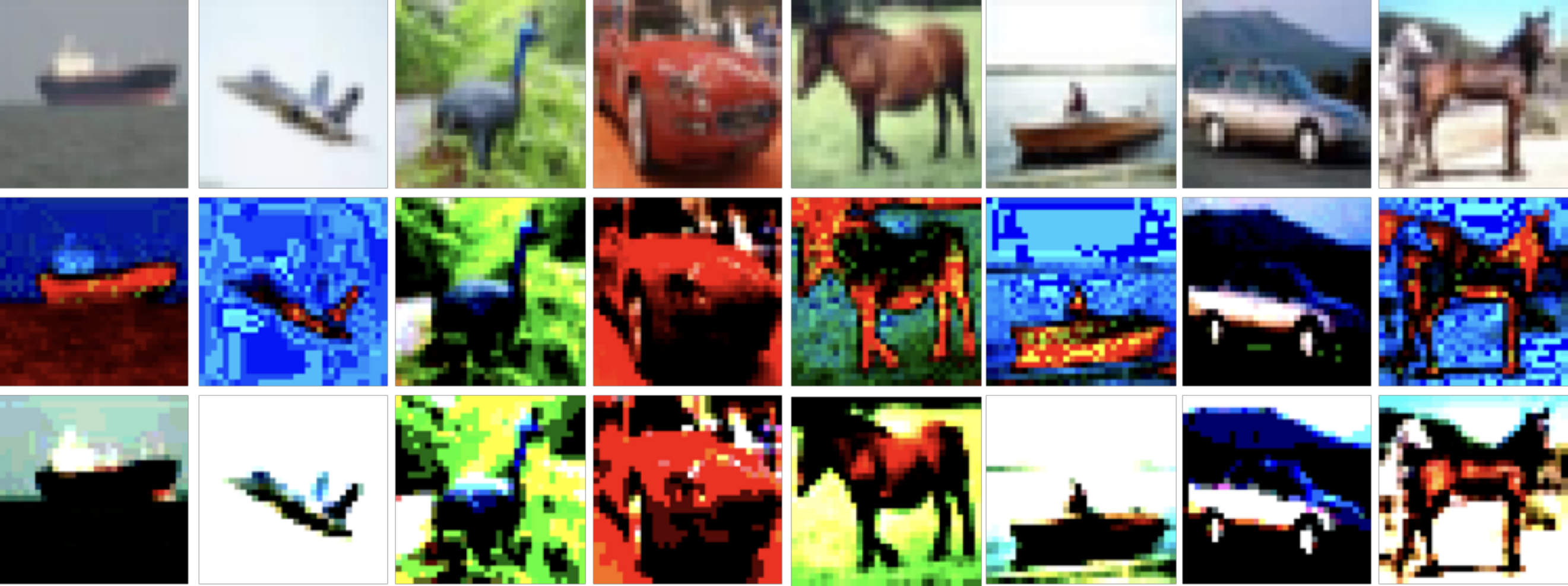}}
    \caption{Visualization of CIFAR-10 Images. First row: visualization of the CIFAR-10 images coded by the RGB scheme; second row: visualization of the images coded by 1-dimension full lookup tables; third row: visualization of the images coded by compressed lookup tables with CMP-Rate 5.}
    \label{f10}
\end{figure*}

\subsection{Visualization}
In this part, we visualize lookup tables through showing images in the way larger pixel values are visualized as higher color intensity.
Figure \ref{f10} shows eight CIFAR-10 images when they are represented in the original RGB space, 1-dimension full lookup tables, and compressed lookup tables with CMP-Rate 5.
The full lookup tables and compressed lookup tables are learned with VGG-16 on the CIFAR-10 classification task.
Suppose the human visual system prefers the images coded by RGB as shown in the first row of Figure \ref{f10}.
We observe that the code scheme that the DNN favors for CIFAR-10 classification task (i.e., the second row and the third row of Figure \ref{f10}) is different from what the human visual system prefers.
However, it is reasonable because the DNN as a extremely complex function may carry out a task from a perspective that is different from that of humans.

\section{Conclusion and Future Work}
As visual data are almost always represented in a manually designed coding scheme when they are input to DNNs, we have explored whether the inputs to DNNs can be optimally learned end-to-end, and have proposed the paradigm of {\em deep collective learning} which aims to learn the weights of DNNs and the inputs to DNNs simultaneously.
Due to the lack of the research on deep collective learning in computer vision, we have proposed Lookup-VNets as a solution.
Lookup-VNets enable DNNs to learn the optimal inputs automatically for given tasks.
From the perspective of image coding, Lookup-VNets can be considered as learning the optimal image coding scheme automatically for given goals.
Additionally, we have explored various aspects of deep collective learning in computer vision with Lookup-VNets through extensive experiments on four benchmark datasets, i.e., CIFAR-10, CIFAR-100, Tiny ImageNet, and Imagenet.
The experiments have shown several surprising characteristics of Lookup-VNets: (1) the vector dimensions in lookup tables has no influence on the test performance (generalization ability) of Lookup-VNets; (2) the commonly used color space can be compressed up to 4096 times without accuracy dropping on CIFAR-10, and 3375 times on CIFAR-100 and Tiny ImageNet.
We also observe that Lookup-VNets are able to match the performances of the standard DNNs on small datasets and achieve superior performances on large-scale and challenging datasets like ImageNet since large and complex datasets can fully take advantages of the flexible and optimally learned inputs.

\par

Besides the basic aspects of Lookup-VNets studied in this paper, various other aspects can be explored, one of which is to design an effective regularizer with regard to lookup tables for Lookup-VNets. 
It is widely believed that the generalization abilities of DNNs are tightly connected to the stability which can be described as a partial derivative of the output to the input.
In Lookup-VNets, the inputs to DNNs are significantly related to the learned lookup tables.
Thus, developing an appropriate regularizer on lookup tables is intuitively promising to further improve the performances of Lookup-VNets.
We leave this to the future work.

\ifCLASSOPTIONcaptionsoff
  \newpage
\fi



\bibliographystyle{IEEEtran}
\bibliography{lookupvnets.bib}

\begin{thebibliography}{10}
\providecommand{\url}[1]{#1}
\csname url@samestyle\endcsname
\providecommand{\newblock}{\relax}
\providecommand{\bibinfo}[2]{#2}
\providecommand{\BIBentrySTDinterwordspacing}{\spaceskip=0pt\relax}
\providecommand{\BIBentryALTinterwordstretchfactor}{4}
\providecommand{\BIBentryALTinterwordspacing}{\spaceskip=\fontdimen2\font plus
\BIBentryALTinterwordstretchfactor\fontdimen3\font minus
  \fontdimen4\font\relax}
\providecommand{\BIBforeignlanguage}[2]{{%
\expandafter\ifx\csname l@#1\endcsname\relax
\typeout{** WARNING: IEEEtran.bst: No hyphenation pattern has been}%
\typeout{** loaded for the language `#1'. Using the pattern for}%
\typeout{** the default language instead.}%
\else
\language=\csname l@#1\endcsname
\fi
#2}}
\providecommand{\BIBdecl}{\relax}
\BIBdecl

\bibitem{krizhevsky2012imagenet}
A.~Krizhevsky, I.~Sutskever, and G.~E. Hinton, ``Imagenet classification with
  deep convolutional neural networks,'' in \emph{Advances in neural information
  processing systems}, 2012, pp. 1097--1105.

\bibitem{long2015fully}
J.~Long, E.~Shelhamer, and T.~Darrell, ``Fully convolutional networks for
  semantic segmentation,'' in \emph{Proceedings of the IEEE conference on
  computer vision and pattern recognition}, 2015, pp. 3431--3440.

\bibitem{redmon2016you}
J.~Redmon, S.~Divvala, R.~Girshick, and A.~Farhadi, ``You only look once:
  Unified, real-time object detection,'' in \emph{Proceedings of the IEEE
  conference on computer vision and pattern recognition}, 2016, pp. 779--788.

\bibitem{mobahi2009deep}
H.~Mobahi, R.~Collobert, and J.~Weston, ``Deep learning from temporal coherence
  in video,'' in \emph{Proceedings of the 26th Annual International Conference
  on Machine Learning}, 2009, pp. 737--744.

\bibitem{he2016deep}
K.~He, X.~Zhang, S.~Ren, and J.~Sun, ``Deep residual learning for image
  recognition,'' in \emph{Proceedings of the IEEE conference on computer vision
  and pattern recognition}, 2016, pp. 770--778.

\bibitem{ren2015faster}
S.~Ren, K.~He, R.~Girshick, and J.~Sun, ``Faster r-cnn: Towards real-time
  object detection with region proposal networks,'' in \emph{Advances in neural
  information processing systems}, 2015, pp. 91--99.

\bibitem{liu2016ssd}
W.~Liu, D.~Anguelov, D.~Erhan, C.~Szegedy, S.~Reed, C.-Y. Fu, and A.~C. Berg,
  ``Ssd: Single shot multibox detector,'' in \emph{European conference on
  computer vision}.\hskip 1em plus 0.5em minus 0.4em\relax Springer, 2016, pp.
  21--37.

\bibitem{mistry2016micro}
K.~Mistry, L.~Zhang, S.~C. Neoh, C.~P. Lim, and B.~Fielding, ``A micro-ga
  embedded pso feature selection approach to intelligent facial emotion
  recognition,'' \emph{IEEE transactions on cybernetics}, vol.~47, no.~6, pp.
  1496--1509, 2016.

\bibitem{zhang2018multiview}
C.~Zhang, J.~Cheng, and Q.~Tian, ``Multiview semantic representation for visual
  recognition,'' \emph{IEEE transactions on cybernetics}, 2018.

\bibitem{lan2018prior}
R.~Lan, Y.~Zhou, Z.~Liu, and X.~Luo, ``Prior knowledge-based probabilistic
  collaborative representation for visual recognition,'' \emph{IEEE
  transactions on cybernetics}, 2018.

\bibitem{hornik1989multilayer}
K.~Hornik, M.~Stinchcombe, H.~White \emph{et~al.}, ``Multilayer feedforward
  networks are universal approximators.'' \emph{Neural networks}, vol.~2,
  no.~5, pp. 359--366, 1989.

\bibitem{leshno1993multilayer}
M.~Leshno, V.~Y. Lin, A.~Pinkus, and S.~Schocken, ``Multilayer feedforward
  networks with a nonpolynomial activation function can approximate any
  function,'' \emph{Neural networks}, vol.~6, no.~6, pp. 861--867, 1993.

\bibitem{Simonyan15}
K.~Simonyan and A.~Zisserman, ``Very deep convolutional networks for
  large-scale image recognition,'' in \emph{International Conference on
  Learning Representations}, 2015.

\bibitem{agostinelli2014learning}
F.~Agostinelli, M.~Hoffman, P.~Sadowski, and P.~Baldi, ``Learning activation
  functions to improve deep neural networks,'' in \emph{International
  Conference on Learning Representations}, 2015.

\bibitem{molina2019pad}
A.~Molina, P.~Schramowski, and K.~Kersting, ``Pad$\backslash$'e activation
  units: End-to-end learning of flexible activation functions in deep
  networks,'' in \emph{International Conference on Learning Representation},
  2020.

\bibitem{lee2016generalizing}
C.-Y. Lee, P.~W. Gallagher, and Z.~Tu, ``Generalizing pooling functions in
  convolutional neural networks: Mixed, gated, and tree,'' in \emph{Artificial
  intelligence and statistics}, 2016, pp. 464--472.

\bibitem{streeter2019learning}
M.~Streeter, ``Learning optimal linear regularizers,'' \emph{Proceedings of the
  36th International Conference on Machine Learning, PMLR 97:5996-6004}, 2019.

\bibitem{bengio2003neural}
Y.~Bengio, R.~Ducharme, P.~Vincent, and C.~Jauvin, ``A neural probabilistic
  language model,'' \emph{Journal of machine learning research}, vol.~3, no.
  Feb, pp. 1137--1155, 2003.

\bibitem{ouyang2015sentiment}
X.~Ouyang, P.~Zhou, C.~H. Li, and L.~Liu, ``Sentiment analysis using
  convolutional neural network,'' in \emph{2015 IEEE International Conference
  on Computer and Information Technology; Ubiquitous Computing and
  Communications; Dependable, Autonomic and Secure Computing; Pervasive
  Intelligence and Computing}.\hskip 1em plus 0.5em minus 0.4em\relax IEEE,
  2015, pp. 2359--2364.

\bibitem{zhang2018alignment}
B.~Zhang, D.~Xiong, J.~Su, and Y.~Qin, ``Alignment-supervised bidimensional
  attention-based recursive autoencoders for bilingual phrase representation,''
  \emph{IEEE transactions on cybernetics}, vol.~50, no.~2, pp. 503--513, 2018.

\bibitem{krizhevsky2009learning}
A.~Krizhevsky, G.~Hinton \emph{et~al.}, ``Learning multiple layers of features
  from tiny images,'' Citeseer, Tech. Rep., 2009.

\bibitem{deng2009imagenet}
J.~Deng, W.~Dong, R.~Socher, L.-J. Li, K.~Li, and L.~Fei-Fei, ``Imagenet: A
  large-scale hierarchical image database,'' in \emph{2009 IEEE conference on
  computer vision and pattern recognition}.\hskip 1em plus 0.5em minus
  0.4em\relax Ieee, 2009, pp. 248--255.

\bibitem{lee2017fully}
J.~Lee, K.~Cho, and T.~Hofmann, ``Fully character-level neural machine
  translation without explicit segmentation,'' \emph{Transactions of the
  Association for Computational Linguistics}, vol.~5, pp. 365--378, 2017.

\bibitem{sennrich-etal-2016-neural}
R.~Sennrich, B.~Haddow, and A.~Birch, ``Neural machine translation of rare
  words with subword units,'' in \emph{Proceedings of the 54th Annual Meeting
  of the Association for Computational Linguistics (Volume 1: Long
  Papers)}.\hskip 1em plus 0.5em minus 0.4em\relax Berlin, Germany: Association
  for Computational Linguistics, Aug. 2016, pp. 1715--1725.

\bibitem{wu2016google}
Y.~Wu, M.~Schuster, Z.~Chen, Q.~V. Le, M.~Norouzi, W.~Macherey, M.~Krikun,
  Y.~Cao, Q.~Gao, K.~Macherey \emph{et~al.}, ``Google's neural machine
  translation system: Bridging the gap between human and machine translation,''
  2016.

\bibitem{mikolov2013efficient}
T.~Mikolov, K.~Chen, G.~Corrado, and J.~Dean, ``Efficient estimation of word
  representations in vector space,'' in \emph{International Conference on
  Learning Representation}, 2013.

\bibitem{pennington2014glove}
J.~Pennington, R.~Socher, and C.~D. Manning, ``Glove: Global vectors for word
  representation,'' in \emph{Proceedings of the 2014 conference on empirical
  methods in natural language processing (EMNLP)}, 2014, pp. 1532--1543.

\bibitem{baroni2014don}
M.~Baroni, G.~Dinu, and G.~Kruszewski, ``Don’t count, predict! a systematic
  comparison of context-counting vs. context-predicting semantic vectors,'' in
  \emph{Proceedings of the 52nd Annual Meeting of the Association for
  Computational Linguistics (Volume 1: Long Papers)}, 2014, pp. 238--247.

\bibitem{hochreiter1997long}
S.~Hochreiter and J.~Schmidhuber, ``Long short-term memory,'' \emph{Neural
  computation}, vol.~9, no.~8, pp. 1735--1780, 1997.

\bibitem{zhang2015character}
X.~Zhang, J.~Zhao, and Y.~LeCun, ``Character-level convolutional networks for
  text classification,'' in \emph{Advances in neural information processing
  systems}, 2015, pp. 649--657.

\bibitem{conneau2016very}
A.~Conneau, H.~Schwenk, L.~Barrault, and Y.~Lecun, ``Very deep convolutional
  networks for text classification,'' \emph{arXiv preprint arXiv:1606.01781},
  2016.

\bibitem{yao2019graph}
L.~Yao, C.~Mao, and Y.~Luo, ``Graph convolutional networks for text
  classification,'' in \emph{Proceedings of the AAAI Conference on Artificial
  Intelligence}, vol.~33, 2019, pp. 7370--7377.

\bibitem{he2015delving}
K.~He, X.~Zhang, S.~Ren, and J.~Sun, ``Delving deep into rectifiers: Surpassing
  human-level performance on imagenet classification,'' in \emph{Proceedings of
  the IEEE international conference on computer vision}, 2015, pp. 1026--1034.

\bibitem{lin2014learning}
D.~Lin, C.~Lu, R.~Liao, and J.~Jia, ``Learning important spatial pooling
  regions for scene classification,'' in \emph{Proceedings of the IEEE
  Conference on Computer Vision and Pattern Recognition}, 2014, pp. 3726--3733.

\bibitem{zhu2019empirical}
X.~Zhu, D.~Cheng, Z.~Zhang, S.~Lin, and J.~Dai, ``An empirical study of spatial
  attention mechanisms in deep networks,'' in \emph{Proceedings of the IEEE
  International Conference on Computer Vision}, 2019, pp. 6688--6697.

\bibitem{sun2017learning}
M.~Sun, Z.~Song, X.~Jiang, J.~Pan, and Y.~Pang, ``Learning pooling for
  convolutional neural network,'' \emph{Neurocomputing}, vol. 224, pp. 96--104,
  2017.

\bibitem{kingma2014adam}
D.~P. Kingma and J.~Ba, ``Adam: A method for stochastic optimization,''
  \emph{the 3rd International Conference for Learning Representations}, 2015.

\bibitem{szegedy2015going}
C.~Szegedy, W.~Liu, Y.~Jia, P.~Sermanet, S.~Reed, D.~Anguelov, D.~Erhan,
  V.~Vanhoucke, and A.~Rabinovich, ``Going deeper with convolutions,'' in
  \emph{Proceedings of the IEEE conference on computer vision and pattern
  recognition}, 2015, pp. 1--9.

\bibitem{Zagoruyko2016WRN}
S.~Zagoruyko and N.~Komodakis, ``Wide residual networks,'' in \emph{BMVC},
  2016.

\bibitem{liu2015very}
S.~Liu and W.~Deng, ``Very deep convolutional neural network based image
  classification using small training sample size,'' in \emph{2015 3rd IAPR
  Asian conference on pattern recognition (ACPR)}.\hskip 1em plus 0.5em minus
  0.4em\relax IEEE, 2015, pp. 730--734.

\end{thebibliography}
\end{document}